%% file: main.tex
\documentclass{article}

\usepackage[final]{corl_2026} 

\usepackage{enumitem}
\usepackage{algorithm}
\usepackage{algpseudocode}
\usepackage{amsmath}

\usepackage{amssymb}

\usepackage{booktabs}
\usepackage{pifont}
\usepackage{makecell}
\usepackage{array}
\usepackage{colortbl}
\usepackage{caption}

\usepackage{enumitem}

\usepackage[normalem]{ulem}
\usepackage[usenames,dvipsnames,table]{xcolor}
\hypersetup{
    colorlinks=true,
    linkcolor=MidnightBlue,
    filecolor=MidnightBlue,      
    urlcolor=MidnightBlue,
    citecolor=MidnightBlue,
}
\usepackage{cleveref}

\usepackage{graphicx}
\usepackage{wrapfig}
\usepackage{duckuments}
\usepackage{todonotes}

\definecolor{stage2row}{RGB}{232, 248, 232} 
\definecolor{stage3row}{RGB}{230, 240, 255} 
\definecolor{stage2hdr}{RGB}{46, 139, 87} 
\definecolor{stage3hdr}{RGB}{30, 90, 170} 

\newcommand{\cmark}{{\footnotesize\ding{51}}}
\newcommand{\xmark}{\textcolor{gray}{\footnotesize\ding{55}}}

\renewcommand{\arraystretch}{1.08}

\title{ADEPT: \uline{A}ccelerating \uline{De}xterity via \uline{P}re-\uline{T}raining and Post-Training using Reinforcement Learning}

\author{
  \textbf{Jayjun Lee$^\dagger$$^\ddagger$ Jessica Yin$^\dagger$ Asif Rana$^\dagger$ Nicholas Blauch$^\dagger$ Octi Zhang$^\dagger$ Sam Mady$^\dagger$}\\
  \textbf{Mohak Bhardwaj$^\dagger$ Nima Fazeli$^\ddagger$ Nathan Ratliff$^\dagger$ Karl Van Wyk$^\dagger$ Ankur Handa$^\dagger$}\\
  $^\dagger$NVIDIA Corporation\\
  $^\ddagger$Robotics Department, University of Michigan\\[5pt]
  \textbf{\textcolor{MidnightBlue}{\href{https://adept-dexterity.github.io}{adept-dexterity.github.io}}}
}

\begin{document}
\maketitle

\input{sections/0_abstract}
\input{sections/1_introduction}

\input{sections/2_related_works}
\input{sections/3_methodology}

\input{sections/4_experiments}

\input{sections/6_limitations}
\input{sections/5_discussion}

\clearpage


\acknowledgments{
We thank Kelly Guo and Gavriel State for their help with simulation in Isaac Lab, and Alex Lefort for assistance with 3D printing and the experimental setup.
}
\bibliography{references} 

\input{sections/appendix}

\end{document}

%% file: sections/0_abstract.tex
\vspace{-8mm}
\begin{figure}[h]
\centering
  \includegraphics[width=1.0\textwidth]{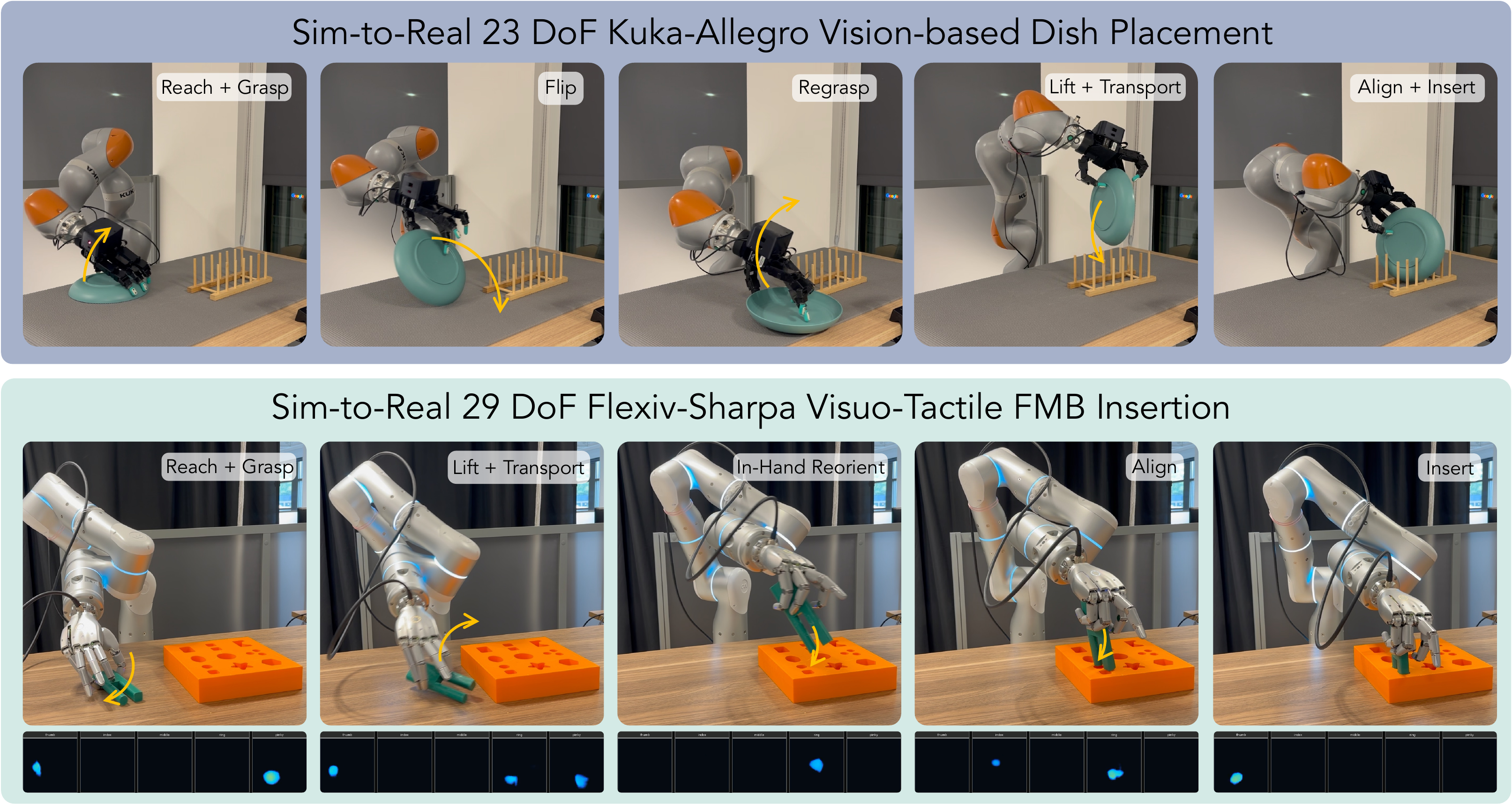}
  \caption{ADEPT enables sim-to-real RL of high DoF arm-hand robots to reach, grasp, lift, reorient, transport, align, and insert. We demonstrate on a 23 DoF Kuka-Allegro from two RGB cameras and on a 29 DoF Flexiv-Sharpa from two RGB cameras and five fingertip tactile sensors.}
\end{figure}

\begin{abstract}
We introduce \uline{A}ccelerating \uline{DE}xterity via \uline{P}re-\uline{T}raining (ADEPT), a large-scale reinforcement learning (RL) framework for learning sim-to-real transferable dexterity across high degree-of-freedom (DoF) robot embodiments that can solve long-horizon tasks directly from raw visuo-tactile perception. 
ADEPT pretrains a dexterous policy on a generic object reposing task, then post-trains downstream policies with this pretrained behavior as a prior. ADEPT enables learning new behaviors that are otherwise difficult to discover from scratch on multi-fingered robots and avoids learning the same set of skills over again for every new downstream task.
The pretrained policy zero-shots the reposing phase of downstream tasks, but na\"ive RL fine-tuning rapidly degrades this capability during transfer. 
We address this with a stable post-training recipe combining behavior-cloning distillation, critic warm-up, and conservative on-policy updates. 
To safely exploit the full kinematic dexterity, we introduce a joint-space Geometric Fabric that mediates between the RL policy and the robot.  
We distill post-trained teachers into perceptive students that zero-shot sim-to-real transfer on two embodiments: a 23 DoF Kuka-Allegro with two RGB cameras, and a 29 DoF Flexiv-Sharpa with two RGB cameras and five vision-based tactile sensors, and can solve long-horizon tasks from challenging initial states with dexterity at human-level speed. 
\end{abstract}

\keywords{Multisensory Dexterity, Sim-to-Real Reinforcement Learning} 

%% file: sections/1_introduction.tex


\section{Introduction}

Dexterous manipulation with high degree-of-freedom (DoF) robotic systems, such as arms equipped with anthropomorphic multi-fingered hands, remains a major challenge for robot learning~\cite{solvingrubikscube, dextreme, dextrahg, visualdexterity}. Multi-fingered arm-hand systems combine high-dimensional state and action spaces with contact-rich interactions, making useful behaviors hard to discover from sparse task-specific rewards. Even with massively parallel GPU simulation~\cite{isaacgym}, RL policies trained from scratch for one task rarely transfer to another; each new task starts over, rediscovering the same low-level skills like reaching, grasping, lifting, and reorientation before any task-specific behavior can emerge.

This motivates a different paradigm. Rather than training an RL specialist from scratch for each new dexterous task, we propose first acquiring a general-purpose dexterous foundation and then adapting it for specific downstream tasks~\cite{urlb, calql}. Generic pretraining supplies reusable manipulation skills that can be adapted for downstream learning~\cite{ewc}.


We introduce \uline{A}ccelerating \uline{DE}xterity via \uline{P}re-\uline{T}raining (ADEPT), a framework that pretrains foundational arm-hand-object dexterity on a generic object reposing task \cite{dexpbt} and post-trains it for downstream contact-rich tasks. We find that this pretrained policy zero-shots the reposing segment of downstream tasks despite never having seen them during training, but transferring the policy and fast adaptation to a contact-rich downstream task is non-trivial. Na\"ive RL fine-tuning often collapses the pretrained behaviors before they can be specialized to the new task~\cite{rlsrazor}. We trace this failure to mismatched observation spaces, misaligned value estimates, and excessive policy drift during early PPO~\cite{ppo} updates, and address it with a structured post-training procedure of actor distillation, critic warm-up, and conservative on-policy updates, respectively. 


Mirroring teacher training, a two-stage distillation curriculum transfers pretrained reposing knowledge, then post-trained task knowledge, into end-to-end perceptive students for zero-shot deployment on two arm-hand embodiments. On a 23 DoF Kuka iiwa7 + Allegro system, an RGB student solves the Functional Manipulation Benchmark (FMB)~\cite{fmb} peg insertion task end-to-end as a single learned policy on two peg geometries that span the benchmark's difficulty range: a symmetric star peg and the asymmetric square-and-round peg, the most challenging geometry in FMB, and additionally solves dish-rack placement. On a 29 DoF Flexiv-Sharpa system, a visuo-tactile student solves FMB peg insertion from two RGB cameras and five fingertip tactile sensors. FMB was originally set up for parallel-jaw grippers, which require external fixtures to reorient and multiple stages of interaction with the object, both of which our approach eschews by using multi-fingered hands while discovering in-hand dexterous manipulations not easily possible with human demonstrations. To summarize, our contributions are:
\begin{itemize}[leftmargin=*]
    \item \textbf{ADEPT}, a framework for pretraining foundational arm--hand--object dexterity on a generic reposing task using RL and post-training it for contact-rich downstream tasks that accelerates and makes full DoF dexterous state-based RL tractable.
    \item A novel \textbf{full joint configuration space (Cspace) geometric fabric}~\cite{geometricfabrics} that gives the RL policy access to the full kinematic dexterity of high-DoF arm--hand systems while enforcing hardware safety (preventing collisions and hitting joint limits), in contrast to prior fabric-guided dexterous policies~\cite{dextrahg, dextrahrgb} that restrict the hand to a low-dimensional PCA grasp subspace.
    \item A \textbf{structured RL post-training recipe} and \textbf{distillation curriculum} that together enable zero-shot sim-to-real on contact-rich, long-horizon dexterous insertion tasks across multiple arm-hand embodiments. To our knowledge, our results represent the \textit{first demonstration} of challenging pick-reorient-insert with robotic arm-hand systems via sim-to-real RL from raw perception without demonstrations or pose trackers, on a vision-based Kuka-Allegro and a visuo-tactile Flexiv-Sharpa. Each downstream task is trained independently using the same pretrained reposing policy for its embodiment.
    \item A $2\times$--$14\times$ \textbf{execution-time speedup} over the FMB~\cite{fmb} parallel-jaw gripper: our multi-fingered policies solve each task in $5$--$10$\,s versus $20$--$70$\,s per trial with external fixtures and multi-stage regrasp decomposition.
\end{itemize}

%% file: sections/2_related_works.tex
\section{Related Work}
\noindent\textbf{Sim-to-Real Dexterous Manipulation.} Large-scale PPO with domain randomization enabled sim-to-real dexterity~\cite{andrychowicz2020dex, solvingrubikscube} and in-hand reorientation~\cite{dextreme, chen2022reorient, visualdexterity, hora, qi2023generalrotation}. ManipGen~\cite{manipgen} composes local simulation-trained policies with motion planners, while OmniReset~\cite{omnireset} diversifies reset states to elicit dexterity, a form of generality that we instead achieve at the policy level through reposing pretraining. SimToolReal~\cite{simtoolreal} trains a task-agnostic, object-centric RL policy for zero-shot tool use but conditions deployment on estimated object and goal poses. ADEPT instead deploys perception-based students that consume raw RGB and fingertip tactile directly, without a pose estimator in the loop. DextrAH-G and DextrAH-RGB~\cite{dextrahg, dextrahrgb} use geometric fabrics as a low-level controller for sim-to-real grasping. DemoStart~\cite{demostart} learns pixel-based grasp--reorient--insert on a three-fingered hand using demonstrations to seed learning. To our knowledge, prior work has not demonstrated zero-shot sim-to-real RL for long-horizon pick--reorient--insert on a high-DoF arm-hand system from raw visual or visuo-tactile observations without demonstrations or a pose estimator.

\noindent\textbf{Reinforcement Learning Pretraining and Transfer.} RL pretraining and adaptation~\cite{urlb} use behavior priors and regularization to mitigate distribution shift and forgetting during online adaptation~\cite{wsrl, adaptiveBCreg, calql, rlsrazor, ewc}. DexGen~\cite{dexgen} learns a diffusion controller from RL-generated in-hand grasp transitions to refine coarse teleoperation commands. Contemporaneous Play2Perfect~\cite{play2perfect} finetunes a dexterous play prior for assembly and deploys state-based teacher policies with externally estimated object poses and supplied goal poses. ADEPT pretrains coordinated arm-hand skills for reaching, grasping, lifting, reorienting, and transporting tabletop objects. Its structured RL post-training addresses changes in observation space and reward under task transfer through actor distillation, frozen-actor critic warm-up, and conservative PPO updates. Our resulting vision and visuo-tactile students deploy without an external pose estimator or explicit goal pose inputs. 

\noindent\textbf{Geometric Fabrics for Policy Learning.} Geometric fabrics~\cite{geometricfabrics} provide smooth second-order action priors with stability guarantees and collision and joint-limit avoidance, building on geometrically consistent reactive control~\cite{rmp, rmpflow, optfabrics}.
DextrAH-G and DextrAH-RGB~\cite{dextrahg, dextrahrgb} restrict hand control to a 5D PCA grasp subspace plus a 6D palm pose target, limiting finger coordination for contact-rich manipulation.
ADEPT commands the full joint configuration space (Cspace), exposing full arm-hand kinematic dexterity at the cost of a harder learning problem.

\noindent\textbf{Tactile Sensing for Sim-to-Real Dexterous Manipulation.} Recent work bridges simulated and real tactile signals through shared representations. HydroShear~\cite{hydroshear} simulates hydroelastic shear for zero-shot manipulation with parallel-jaw grippers. TacMap~\cite{tacmap} uses geometry-consistent penetration-depth maps, while SaTA~\cite{sata} anchors per-finger features through FiLM conditioning on fingertip positions. ADEPT combines TacMap-derived depth and binary contact maps with SaTA-style anchoring for spatially grounded fingertip contact observations.

%% file: sections/3_methodology.tex
\section{Methodology}

\begin{figure}[h]
\centering
  \includegraphics[width=\textwidth]{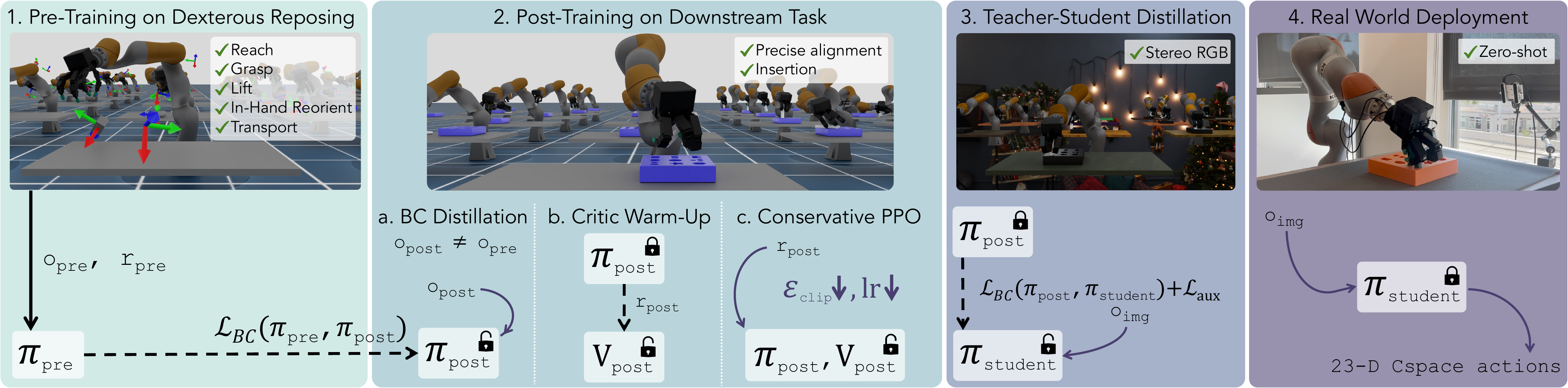}
  \caption{\textbf{ADEPT Overview.} (1) Pre-train $(\pi_{\mathrm{pre}}, V_{\mathrm{pre}})$ via PPO on a generic reposing task. (2) Post-train into $(\pi_{\mathrm{post}}, V_{\mathrm{post}})$ for the downstream contact-rich task via BC distillation, frozen actor critic warm-up, and conservative PPO. (3) Distill $\pi_{\mathrm{post}}$ into a perceptive (vision or visuo-tactile) student $\pi_\text{student}$. (4) Deploy $\pi_\text{student}$ zero-shot on the real robot.}
  \label{fig:adept-overview}
\end{figure}

We propose a framework for pretraining foundational dexterity on a generic task objective and post-training for contact-rich downstream tasks as outlined in Fig.~\ref{fig:adept-overview}. Our approach consists of three stages: (1) pretraining a dexterous policy in simulation on a generic object manipulation task, (2) transferring the pretrained behavior through distillation and adapting through massively parallel on-policy RL post-training while preventing catastrophic forgetting through stable policy updates, and (3) teacher-student distillation using a task specialist teacher distilled into a stereo RGB vision-based student that can be zero-shot deployed in the real-world. 


\subsection{Problem Formulation}
\label{sec:problem_formulation}

We cast dexterous manipulation as a discrete-time Markov decision process (MDP) $\mathcal{M} = (\mathcal{S}, \mathcal{A}, p, r, \gamma)$ with continuous state space $\mathcal{S}$, action space $\mathcal{A} = [-1, 1]^{n_q}$ ($n_q = 23$ for Kuka-Allegro, $29$ for Flexiv-Sharpa), transition dynamics $p$, scalar reward $r$, and discount factor $\gamma$. We learn a stochastic policy $\pi_\theta(a_t \mid o_t)$ that maximizes the expected discounted return $J(\theta) = \mathbb{E}_{\tau \sim \pi_\theta} \left[ \sum_{t=0}^{T} \gamma^t r(s_t, a_t) \right]$ via PPO~\cite{ppo} with an asymmetric actor-critic. ADEPT operates over two MDPs $\mathcal{M}_{\mathrm{pre}}$ and $\mathcal{M}_{\mathrm{post}}$ for the pretraining (Sec.~\ref{sec:pretraining}) and post-training (Sec.~\ref{sec:posttraining}) tasks respectively. They share the action space $\mathcal{A}$ but differ in observation space, dynamics, and reward. $o^{\mathrm{post}}$ adds task-specific receptacle pose $\mathbf{p}_{\mathrm{rec}}$ and contact forces $\mathbf{f}_{\mathrm{or}}$ while removing explicit object pose input from $o^{\mathrm{pre}}$ to reduce the observation gap for sim-to-real transfer (Tabs.~\ref{tab:obs-layout-v2} and
\ref{tab:obs-layout-rizon-sharpa} in Appx.~\ref{appx:obstable}). We first solve $\mathcal{M}_{\mathrm{pre}}$ with PPO to obtain $(\pi_{\mathrm{pre}}, V_{\mathrm{pre}})$, then BC distill $\pi_{\mathrm{pre}}$ into a new downstream actor and learn $(\pi_{\mathrm{post}}, V_{\mathrm{post}})$ on $\mathcal{M}_{\mathrm{post}}$ through structured post-training.

\subsection{Pre-Training Foundational Dexterity}
\label{sec:pretraining}
We pretrain a dexterous teacher actor-critic with PPO on a generic object reposing task designed to instill broad manipulation capabilities. At each episode, one of 16 primitive shapes (cylinders, cuboids, spheres, cones) at a randomized scale is spawned on the table, and the policy must execute a full sequence of reaching, grasping, lifting, in-hand reorientation, transporting, and reposing the object to a sampled target goal pose. Automatic domain randomization (ADR)~\cite{solvingrubikscube, dextreme} advances goals and environmental complexity as success improves, including annealing gravity from $0$ to $-9.81$\,m/s$^2$. Population-Based Training (PBT)~\cite{pbt, dexpbt} searches PPO hyperparameters (Appx.~\ref{appx:pbt}). This setup encourages the emergence of reusable dexterous skills (precise grasps, coordinated finger motion, and in-hand reorientation across varied object geometries and scales), yielding foundational dexterity that solves the reposing segment of downstream tasks in a zero-shot manner (Tab.~\ref{tab:zeroshot_adr_levels}). Refer to Appx.~\ref{appx:reposingmdp} for the full task specification, observation space, and reward function. Note that we use object point clouds to represent objects, which has shown better zero-shot generalization for downstream tasks. At $1.2$ million environment transitions per second and a $60$\,Hz control frequency, approximately $100$ hours of Flexiv-Sharpa pretraining and $25$ hours of Kuka-Allegro pretraining correspond to $228$ and $57$ years of aggregate simulated robot experience, respectively (Appx.~\ref{appx:pbt}).

\subsection{Post-training Dexterous RL Specialists on Downstream Tasks}
\label{sec:posttraining}

\textbf{Challenges in Transferring Pretrained Dexterity.} Direct PPO fine-tuning rapidly degrades the pretrained policy: reposing success falls to $0\%$ within a few updates (Fig.~\ref{fig:post-training}, inset; Sec.~\ref{sec:sim-exp}). This collapse occurs even from a policy with strong zero-shot performance, indicating that the failure arises during training rather than from insufficient pretraining. We attribute this to a number of mismatches under the transfer learning setup: (1) change in task reward function, (2) extra observations that are only available in the downstream task (e.g. contact forces between object-receptacle and pose of the receptacle), (3) poor value or advantage estimates, and thus (4) large policy updates.

\textbf{Structured RL Adaptation.} We frame downstream learning as layering new behaviors on top of the pretrained dexterous prior rather than relearning manipulation from scratch: each iteration of on-policy PPO is restricted to a local region of the pretrained policy space, so the policy retains core reaching, grasping, lifting, and reorientation skills while incorporating downstream task-specific observations $\mathbf{p}_{\mathrm{rec}}$ and $\mathbf{f}_{\mathrm{or}}$ and reward. Actor distillation transfers behavior to the new observation space, critic warm-up recalibrates value estimates, and conservative policy updates still allow downstream behavioral adaptation. 
Concretely, we transfer the pretrained actor--critic $(\pi_{\mathrm{pre}}, V_{\mathrm{pre}})$ into the downstream task in three steps. \textbf{(1) BC actor distillation.} We distill $\pi_{\mathrm{pre}}$ into a new downstream actor $\pi_\text{post}$ with the downstream observation space using supervised imitation for 40k iterations. \textbf{(2) Critic warm-up.} With $\pi_\text{post}$ frozen, we train a fresh critic $V_\text{post}$ on its rollouts under the downstream reward for 20 PPO iterations ($\sim1$M environment steps per GPU with 4096 environments) before updating the actor. \textbf{(3) Conservative PPO.} We finally unfreeze $\pi_\text{post}$ and jointly update $(\pi_\text{post}, V_\text{post})$ with PPO using conservative updates 
(actor LR 1e-3 $\to$ 1e-5 with linear decay, PPO clip $\epsilon$: $0.2 \to 0.05$, critic LR fixed at 5e-5).
BC distillation (1) runs against the single ADR 20 goal pose on the goal path illustrated in Fig.~\ref{fig:fmb_task_layout} by grey spheres; critic warm-up (2) and conservative PPO (3) then train directly against the final insertion goal at ADR 50 while the remaining ADR randomizations anneal from level 20; training from scratch must follow the same curriculum from ADR 0. Controlled ablations (Tab.~\ref{tab:posttrain_ablation}) show that the reduced actor learning rate is what prevents collapse, while BC and critic warm-up improve success, stability, and adaptation speed; the tightened clip has little effect. Refer to Alg.~\ref{alg:bootstrap} for the full algorithm and Tab.~\ref{tab:training_hparams} in Appx.~\ref{appx:bootstrapalgorithms} for the training hyperparameters.

\subsection{Teacher-Student Distillation}
\label{sec:distillation}

\textbf{Setup.} After RL post-training yields a state-based teacher for a contact-rich downstream task, we distill it into a deployable student that consumes only proprioception, fabric state, two RGB images, and, on the Flexiv-Sharpa embodiment, five per-finger tactile maps for zero-shot deployment (Tab.~\ref{tab:obs-layout-v2}, Appx.~\ref{appx:obstable}). With DAgger~\cite{dagger}, the teacher provides per-step action targets (mean and variance) over rollouts collected from the student in the same task environment, and the student uses an initially frozen pretrained ResNet backbone with a DextrAH-RGB-style~\cite{dextrahrgb} unfreeze schedule.

\textbf{Auxiliary loss.} Precise insertion requires reliable perception of the peg's orientation, which action cloning does not explicitly supervise. Extending DextrAH-RGB's object-position loss~\cite{dextrahrgb}, we supervise a pose head that predicts the eight corners of the peg's tight bounding box from shared stereo features using simulation ground truth. The objective is $\mathcal{L} = \mathcal{L}_{\mathrm{BC}} + \mathcal{L}_{\mathrm{aux}}$ (Appx.~\ref{appx:distill_loss}).

\textbf{Two-stage student curriculum.} Mirroring teacher pretraining and post-training, we use two distillation stages. (1) \textbf{Student pretraining via the reposing teacher.} We re-task the pretrained reposing teacher (Sec.~\ref{sec:pretraining}) to perform a perception-heavy surrogate of the downstream task: lifting and reorienting the peg to an upright pose at a fixed receptacle location. Distilled into the stereo RGB student, this teacher supplies grasping, lifting, and reorientation behavior, while $\mathcal{L}_{\mathrm{aux}}$ supervises the eight tight-bounding-box corners for peg pose perception. (2) \textbf{Distilling downstream task knowledge.} We initialize downstream distillation from the Stage-1 checkpoint and continue training against the post-trained insertion teacher, so contact-rich refinement builds on the student's learned perception and reposing skills.

\textbf{Tactile perception for visuo-tactile students.} On Flexiv-Sharpa, the student receives images from all five fingertip tactile sensors. TacMap~\cite{tacmap} represents both simulated and calibrated real-sensor outputs as geometry-consistent penetration-depth maps, avoiding additional sim-to-real adaptation. From each depth map we derive a binary contact map by thresholding, encode both channels with a per-finger CNN shared across fingers, and spatially anchor the resulting features via SaTA-style~\cite{sata} FiLM conditioning on the fingertip position obtained via forward kinematics. The per-finger tactile embeddings are then fused with the stereo RGB and proprioception latents, giving the policy a spatially grounded view of hand-object contact. Full architecture and dimensions are in Appx.~\ref{appx:visuotactile_arch}.

\textbf{Randomization.} Both distillation stages use aggressive domain randomization on physics, object perturbations via random wrenches, visual conditions (lighting, background, camera intrinsics and pose), and sensor noise on proprioception and contact signals. Refer to Appx.~\ref{appx:distill_randomization} for more details.

%
%

\subsection{Joint Configuration Space Geometric Fabrics}
\label{sec:cspace_fabrics}


We place a geometric fabric~\cite{geometricfabrics} between the policy and the robot to enforce hardware constraints (joint limits, self- and environmental collision), provide a smooth second-order action prior for exploration, and use a shared low-level controller for simulation and deployment. It is a second-order system on configuration space, $\mathbf{M}_f(\mathbf{q}_f, \dot{\mathbf{q}}_f)\,\ddot{\mathbf{q}}_f + \mathbf{f}_f(\mathbf{q}_f, \dot{\mathbf{q}}_f) + \mathbf{f}_\pi(\mathbf{a}) = \mathbf{0}$, where $\mathbf{f}_f$ collects autonomous geometric and dissipative terms (collision and joint-limit repulsion, Cspace damping, speed control) and $\mathbf{f}_\pi(\mathbf{a})$ is the policy-driven forcing term. 
To support behaviors such as in-hand reposing, finger gaiting, and contact-rich manipulation, we drive the fabric in the full joint configuration space: the policy outputs $\mathbf{a}_t \in [-1,1]^{n_q}$ ($n_q = 23$ for Kuka-Allegro, $29$ for Flexiv-Sharpa), interpreted as per-joint relative deltas that map to a Cspace target consumed by $\mathbf{f}_\pi$.
This exposes all arm--hand joints to the policy while retaining the fabric's safety and stability guarantees~\cite{geometricfabrics}. The same fabric instance runs in simulation and on the real robot, reducing the controller gap. Full fabric components, the relative-target mapping, and the integration loop are deferred to Appx.~\ref{appx:fabric_setup}.

%% file: sections/4_experiments.tex
\section{Experiments}


\subsection{Simulation Experiments}
\label{sec:sim-exp}

\textbf{How well does the pretrained teacher generalize to unseen objects?} We evaluate the pretrained teacher on the reposing task across three object categories in Tab.~\ref{tab:teacher_generalization}: the 16 in-distribution primitive shapes used during pretraining, the FMB pegs (star and square/round) from the downstream insertion task, and VisDex~\cite{visualdexterity} objects. Despite training only on primitives, the Kuka-Allegro teacher matches or slightly exceeds its in-distribution success rate on both out-of-distribution (OOD) sets, and Flexiv-Sharpa stays close to its in-distribution success rate. 

\textbf{How far can the pretrained teacher zero-shot the downstream task?} Tab.~\ref{tab:zeroshot_adr_levels} evaluates FMB peg insertion without post-training. Goals follow an L-shaped path (Fig.~\ref{fig:fmb_task_layout}, Appx.~\ref{appx:downstreammdp}): horizontal lifting and transport from above the peg to above the hole (ADR 0--25), followed by vertical insertion (ADR 25--50). Zero-shot success stays above $50\%$ through ADR $35$ but falls to $0\%$ (Kuka-Allegro) and $0.70\%$ (Flexiv-Sharpa) at the insertion goal (ADR $50$) where the task introduces contact with a receptacle absent from pretraining. We therefore begin all downstream post-training at ADR 20: BC distillation clones the teacher against the ADR 20 goal pose, and from critic warm-up onward the goal pose is set directly to the final insertion goal at ADR 50, the endpoint of the path, while the remaining ADR randomizations anneal from level 20. At ADR 20, the pretrained policy reliably solves FMB reposing, leaving contact-rich insertion to be learned. For dish placement, reaching transfers but grasping does not: the plate is larger and flatter than the pretraining primitives. Post-training nevertheless learns the full task, as shown below.
\vspace{-4mm}

\begin{table}[H]
\centering
\small
\caption{Episodic success rates (SR) of pre-trained reposing teacher on unseen objects. Mean $\pm$ std across seeds at the final ADR level.}
\label{tab:teacher_generalization}
\setlength{\tabcolsep}{8pt}
\renewcommand{\arraystretch}{1.1}
\begin{tabular}{@{}l ccc@{}}
\toprule
\textbf{Metric} & \textbf{Primitive} & \textbf{FMB Peg} & \textbf{VisDex~\cite{visualdexterity} (Fixed Goal)} \\
\midrule
Number of objects & 16 & 2 & 152 \\
Kuka-Allegro SR $\uparrow$ & $0.73 \pm 0.003$ & $0.76 \pm 0.003$ & $0.77 \pm 0.011$ \\
Flexiv-Sharpa SR $\uparrow$ & $0.64 \pm 0.007$ & $0.58 \pm 0.011$ & $0.61 \pm 0.015$ \\
\bottomrule
\end{tabular}
\end{table}

\vspace{-8mm}

\begin{table}[H]
\centering
\scriptsize 
\caption{Zero-shot success rate of the pretrained reposing teacher on FMB peg reposing across ADR.}
\label{tab:zeroshot_adr_levels}
\setlength{\tabcolsep}{2.5pt}
\renewcommand{\arraystretch}{1.1}
\begin{tabular}{@{}l ccccccccccc@{}}
\toprule
 & \multicolumn{11}{c}{\textbf{Zero-shot Success Rate (\%) at ADR Level across 1024 Episodes}} \\
\cmidrule(l){2-12}
\textbf{Task (Embodiment)} & 0 & 5 & 10 & 15 & 20 & 25 & 30 & 35 & 40 & 45 & 50 \\
\midrule
FMB Peg (Kuka-Allegro) & $98.6$ & $94.1$ & $84.7$ & $78.6$ & $71.9$ & $67.9$ & $58.7$ & $52.5$ & $39.6$ & $18.0$ & $0.00$ \\
FMB Peg (Flexiv-Sharpa) & $94.5$ & $87.1$ & $80.3$ & $76.8$ & $74.9$ & $70.4$ & $66.8$ & $59.2$ & $52.3$ & $22.9$ & $0.70$ \\
\bottomrule
\end{tabular}
\end{table}

\vspace{-5mm}

\begin{wrapfigure}{R}{0.5\textwidth}
  \centering
  \includegraphics[width=0.5\textwidth]{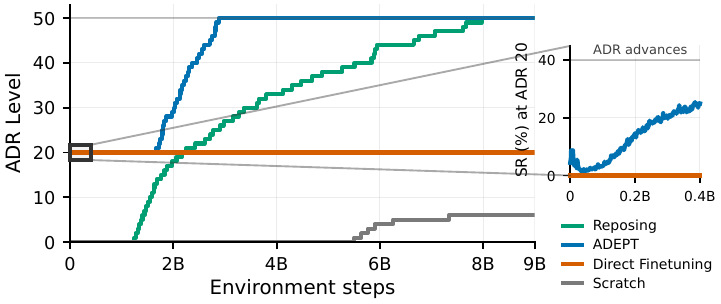}
  \vspace{-1.8em}
  \caption{Training curves on FMB peg insertion comparing \textbf{(1)} training from scratch, \textbf{(2)} direct PPO finetuning of the pretrained reposing teacher, and \textbf{(3)} ADEPT post-training. Blue inset: success-rate collapse for (2) during transfer.}
  \label{fig:post-training}
  \vspace{-0.8em}
\end{wrapfigure}
\textbf{How does ADEPT compare to training from scratch?} Fig.~\ref{fig:post-training} plots ADR level over environment steps for (1) training from scratch, (2) direct PPO finetuning of the pretrained reposing teacher with extra observations, and (3) ADEPT post-training. Fig.~\ref{fig:post-training-wallclock} in Appx.~\ref{appx:training_efficiency} plots the same curves against wall-clock time. ADEPT trains a downstream-task teacher in 3B environment steps on top of the 8B-step reposing pretraining, for a total of 11B steps. Training from scratch is highly seed-sensitive, with most seeds plateauing well below ADR 6 as shown in Fig.~\ref{fig:across-seeds} in Appx.~\ref{appx:training_efficiency} and only a few reaching it after about 9B steps. More importantly, the 8B pretraining cost is amortized across every downstream task, so the marginal cost per new task reduces to the 3B post-training.





\textbf{Why does na\"ive RL finetuning fail?} The blue inset of Fig.~\ref{fig:post-training} shows success at ADR 20, where direct fine-tuning and ADEPT start from the same pretrained teacher. Direct PPO fine-tuning drives success to zero within a few updates, while ADEPT improves. The pretrained critic is calibrated to the reposing reward; under the insertion reward, inaccurate value estimates can produce unreliable advantages and destabilize early actor updates. As the actor drifts, subsequent rollouts move further from the pretrained behavior, leading to rapid forgetting of pretrained skills. ADEPT addresses the observation and reward changes through BC actor distillation and frozen-actor critic warm-up, then reduces the actor learning rate to limit the size of early policy updates and the resulting drift. The ablations below test which changes are needed.

\textbf{Which post-training components matter?} Controlled ablations in Tab.~\ref{tab:posttrain_ablation} decompose the recipe over five independent seeds per variant, all initialized from the same pretrained checkpoint. The reduced actor learning rate is necessary in these ablations: every $10^{-3}$ variant collapses at ADR 20 with zero success, even with BC and critic warm-up in place. A KL penalty $D_{\mathrm{KL}}(\pi_{\mathrm{post}} | \pi_{\mathrm{pre}})$ does not rescue training at $10^{-3}$, and at a learning rate of $10^{-5}$, adding the penalty with $\beta{=}1$ still yields $0\%$ across all five seeds. Given the low learning rate, critic warm-up adds 17.6 percentage points of SR (row c) and BC nearly halves adaptation time ($19.9$ vs.\ $35.2$\,h, row b); removing both prevents every seed from reaching ADR 50. Without BC, the ten new observation dimensions must be injected into the pretrained actor, and we test both constructions: row (g) remaps the actor input layer (nine obsolete inputs removed, ten zero-initialized inputs added), while row (g$^\dagger$) keeps all 391 pretraining inputs and appends the ten. Both stall, so the gap is not an artifact of the remapping. The tightened clip has little effect: loosening it to $0.20$ (row f) achieves similar success and reaches ADR 50 sooner on average. All deployed policies use the tight clip, but these results suggest that the reduced actor learning rate, rather than the tighter clip, is central to stable adaptation.

\begin{table}[h]
\centering
\footnotesize
\setlength{\tabcolsep}{4pt}
\caption{Post-training ablations on KUKA-Allegro FMB peg insertion; $5$ independent non-PBT seeds per row from one pretrained checkpoint. ADR50 counts seeds reaching the final ADR level; time averages those seeds. Train SR is instantaneous training success rate over parallel envs.}
\label{tab:posttrain_ablation}
\scriptsize
\setlength{\tabcolsep}{3.5pt}
\begin{tabular}{@{}l l cc cc c c l@{}}
\toprule
 & \textbf{Variant} & \textbf{BC} & \textbf{WU} & \textbf{LR} & \textbf{Clip} & \textbf{ADR50} & \textbf{Train SR (\%)} & \textbf{Time to ADR 50} \\
\midrule
(a) & ADEPT (full) & \cmark & \cmark & $10^{-5}$ & $.05$ & $5/5$ & $46.0 \pm 1.1$ & $19.9 \pm 1.0$\,h \\
(b) & No BC & \xmark & \cmark & $10^{-5}$ & $.05$ & $4/5$ & $38.5 \pm 1.2$ & $35.2 \pm 4.9$\,h \\
(c) & No warm-up & \cmark & \xmark & $10^{-5}$ & $.05$ & $4/5$ & $28.4 \pm 3.7$ & $20.2 \pm 1.5$\,h \\
(d) & Standard PPO & \cmark & \cmark & $10^{-3}$ & $.20$ & $0/5$ & $0.0 \pm 0.0$ & Collapse (ADR 20) \\
(e) & High LR only & \cmark & \cmark & $10^{-3}$ & $.05$ & $0/5$ & $0.0 \pm 0.0$ & Collapse (ADR 20) \\
(f) & Loose clip & 
 \cmark & \cmark & $10^{-5}$ & $.20$ & $5/5$ & $46.8 \pm 2.0$ & $17.6 \pm 1.8$\,h \\
(g) & No BC/WU (remapped) & \xmark & \xmark & $10^{-5}$ & $.05$ & $0/5$ & $26.6 \pm 7.9$ & Stall (ADR 29--39) \\
(g$^\dagger$) & No BC/WU (append-only) & \xmark & \xmark & $10^{-5}$ & $.05$ & $0/5$ & $27.0 \pm 10.7$ & Stall (ADR 43) \\
(h) & Direct FT & \xmark & \xmark & $10^{-3}$ & $.20$ & $0/5$ & $0.0 \pm 0.0$ & Collapse (ADR 20) \\
(i) & Direct FT + KL & \xmark & \xmark & $10^{-3}$ & $.20$ & $0/5$ & $0.0 \pm 0.0$ & Collapse (ADR 20) \\
\bottomrule
\end{tabular}
\end{table}


\textbf{Emergent natural grasps from ADEPT and refinement from post-training.} Beyond accelerated learning and stable post-training, ADEPT also synthesizes natural grasps. The insertion reward does not constrain how the peg is grasped; both natural and unnatural grasps can satisfy it. From random initialization, PPO maximizes reward, often synthesizing an unnatural grasp in Fig.~\ref{fig:grasp_comparisons}(c), with performance varying across seeds. Reposing pretraining initializes the policy in a region of parameter space where natural grasps emerge. Local post-training updates refine these pretrained natural grasps for the downstream task rather than discovering new grasp modes. Task-misaligned grasps occasionally produced by the reposing teacher, such as bottom grasps at ADR 20, disappear under refinement guided by hand-to-object and object-to-goal alignment rewards in Fig.~\ref{fig:grasp_comparisons}(g-h). Appx.~\ref{appx:qualitative_grasps} compares pretrained, post-trained, and from-scratch grasps.

\textbf{Can we learn novel behaviors through post-training?} If post-training only refined existing behaviors, pretraining coverage would limit its usefulness. Dish placement shows adaptation beyond refinement. Reaching transfers zero-shot, but grasping does not. Unlike the FMB pegs, the plate is larger and flatter than the 16 training primitives (Fig.~\ref{fig:primitive_objects}), and none of the pretrained grasps succeed on it zero-shot. Post-training begins without a competent grasp to refine. Pretraining instead provides a useful starting distribution: reaching and plausible, though unsuccessful, grasp attempts near meaningful contact configurations. From this initialization, downstream RL learns plate grasping and a flip-and-regrasp strategy for upside-down plates, neither successful zero-shot. Whereas FMB adaptation primarily selects and refines existing grasp modes, dish placement shows that the prior can also enable discovery of new behaviors. Tasks with little overlap with pretrained skills may require broader pretraining.

\subsection{Real World Experiments}
\vspace{-3mm}

\begin{figure}[h]
\centering
\includegraphics[width=\textwidth]{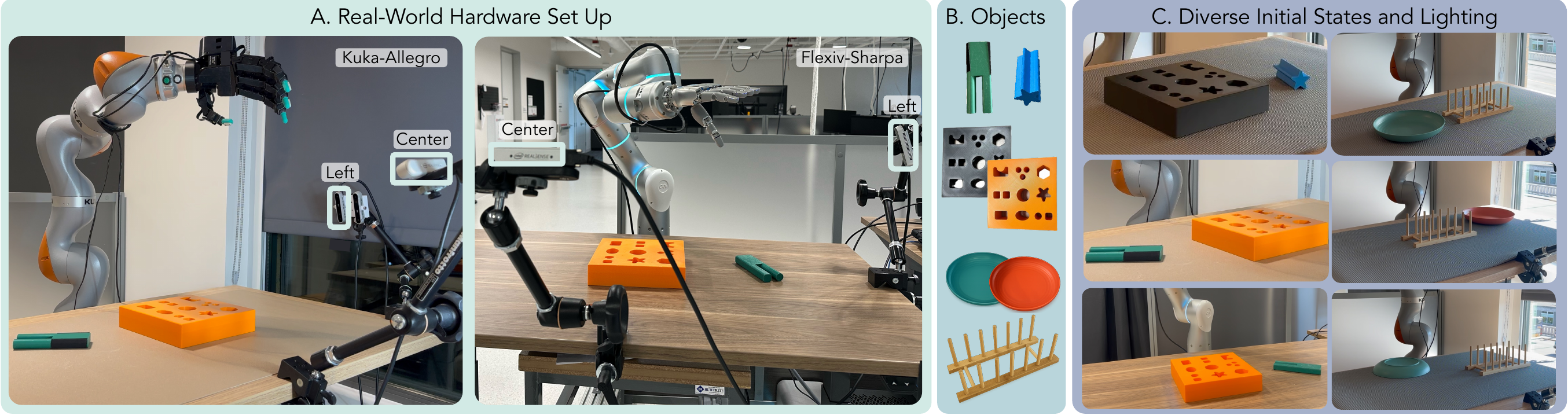}
\caption{Real-world setup. (A) The 23 DoF Kuka-Allegro and 29 DoF Flexiv-Sharpa platforms, each with two RealSense RGB cameras (left and center). (B) FMB pegs and boards, plates, and dish rack. (C) Diverse initial states and lighting used in real-world testing.}
\label{fig:hardware}
\end{figure}

\textbf{Hardware Setup.} We deploy on two arm-hand platforms: a 7 DoF Kuka iiwa7 with a 16 DoF Allegro four-fingered hand ($n_q = 23$), and a 7 DoF Flexiv Rizon with a 22 DoF Sharpa five-fingered hand ($n_q = 29$), each rigidly mounted to a workbench and observed by two calibrated Intel RealSense RGB cameras in the same left-and-center layout (Fig.~\ref{fig:hardware}). Each Sharpa fingertip additionally carries a vision-based tactile sensor whose deformation is exposed to the policy as per-finger TacMap depth maps, the same representation rendered in simulation during training. Deployment details, camera views, and student architecture appear in Appx.~\ref{appx:deployment}, Fig.~\ref{fig:camera_views}, and Appx.~\ref{appx:visuotactile_arch}, respectively.

\textbf{Task Setup.} We evaluate on two FMB~\cite{fmb} pegs in Fig.~\ref{fig:hardware} that span the benchmark's geometric difficulty: a symmetric star peg, whose rotational and top-down symmetry admits multiple valid insertion orientations, and the asymmetric square-and-round peg, the most challenging geometry in FMB, whose square and round legs admit only a single valid insertion orientation. We initialize pegs flat on the table over 30 cm $\times$ 25 cm in spawn range as illustrated in Fig.~\ref{fig:fmb_task_layout} and $[-\pi,\pi]\ \text{rad}$ in yaw and count a trial successful when the peg is fully inserted.

On Kuka-Allegro we additionally evaluate dish-rack placement (Fig.~\ref{fig:hardware}): the policy must grasp a plate, reorient it upright, and seat it in a wooden rack slot. Upside-down plates require a flip-and-regrasp behavior not successfully exhibited by the pretrained policy. Plates are initialized flat on the table, either upright or upside down, over a spawn range similar to the FMB pegs, and a trial counts as successful when the plate is fully seated in a rack slot. In the per-stage breakdown of Tab.~\ref{tab:per_stage_success}, the flip-and-regrasp of upside-down plates is counted as part of the Grasp stage.



\textbf{Results.} 
A single Kuka-Allegro reposing checkpoint post-trains to three downstream tasks, FMB star insertion, FMB square/round insertion, and dish-rack placement, with one post-trained teacher per task and no task-specific pre-training. Zero-shot in the real world, the vision student reaches $5/10$ (star) and $3/10$ (square/round) on FMB and $6/10$ on dish placement; the visuo-tactile Flexiv-Sharpa student reaches $8/10$ on FMB square/round, the only peg trained on that embodiment (Tab.~\ref{tab:per_stage_success}). Post-trained teachers evaluate at $85.0\%$ (Kuka-Allegro, aggregated over both pegs) and $89.2\%$ (Flexiv-Sharpa, square/round) in simulation, and the Kuka-Allegro vision student evaluates at $46.7\%$/$65.2\%$ (square/round, star). These episodic simulation success rates are evaluated over $1024$ episodes at ADR $50$. Training curves report the instantaneous fraction of environments in the success state, which resets after sustained success; this metric differs from episodic evaluation.

\textbf{Star vs.\ square/round.} The star peg's ridges allow fingers to wrap between them for stable grasps, and its symmetry admits multiple valid insertion orientations. The asymmetric peg's round leg yields unstable point contacts, producing the largest per-stage drops at lifting and reorienting (Tab.~\ref{tab:per_stage_success}) and a lower overall success rate.

\textbf{Vision vs.\ visuo-tactile.} On Flexiv-Sharpa, tactile sensing improves real-world success from $3/10$ to $8/10$. The vision-only student often reopens its hand after grasping, drops the object, and repeats grasp attempts, leading to failures during lifting and reorientation (Tab.~\ref{tab:per_stage_success}). These observations suggest ambiguity about whether contact has been established, reflecting limited grasp confidence. With fingertip tactile feedback, it grasps and lifts successfully in every trial and carries that reliability through reorientation, alignment, and insertion. 

\textbf{Distillation curriculum.} The single-stage baseline fails to transfer entirely ($0/10$ on both pegs), while the two-stage curriculum outperforms it by roughly $10\%$ in instantaneous success rates across parallel environments throughout training, supporting the importance of Stage-1 perception pretraining for transfer.

\textbf{Human-level speed.} Our policies solve the entire manipulation sequence as a single continuous behavior at human-comparable timescales ($5$--$10$\,s per trial), reorienting objects in-hand with a single hand. FMB~\cite{fmb} human demonstrations with a parallel-jaw gripper instead decompose the task into grasp, place-on-fixture, regrasp, rotate, and insert primitives, rely on external fixtures, and require $20$--$70$\,s per trial, a $2\times$--$14\times$ execution-time difference.

\vspace{-4mm}

\begin{table}[H]
\centering
\footnotesize
\caption{Per-stage success rates on real-world long-horizon manipulation, trained with ADEPT. Each stage is cumulative: success at stage $k$ requires success at all earlier stages.}
\label{tab:per_stage_success}
\setlength{\tabcolsep}{4pt}
\renewcommand{\arraystretch}{1.1}
\begin{tabular}{@{}lll cccccc@{}}
\toprule
Modality & Robot & Task & Reach & Grasp & Lifting & Reorient & Align & Insert (SR) \\
\midrule
Vision & Kuka-Allegro & FMB Star & 10/10 & 9/10 & 8/10 & 8/10 & 7/10 & 5/10 \\
Vision & Kuka-Allegro & FMB Square/Round & 10/10 & 8/10 & 6/10 & 4/10 & 3/10 & 3/10 \\
Vision & Flexiv-Sharpa & FMB Square/Round & 10/10 & 7/10 & 5/10 & 3/10 & 3/10 & 3/10 \\
Visuo-Tactile & Flexiv-Sharpa & FMB Square/Round & 10/10 & 10/10 & 10/10 & 9/10 & 8/10 & 8/10 \\
Vision & Kuka-Allegro & Dish & 10/10 & 10/10 & 8/10 & 7/10 & 6/10 & 6/10 \\
\bottomrule
\end{tabular}
\end{table}

\vspace{-4mm}

\vspace{-4mm}


%% file: sections/6_limitations.tex
\section{Limitations}
\vspace{-2mm}

Although real-world behavior closely matches simulation, perception remains the primary bottleneck for distillation. Failures frequently coincide with incorrect estimates of the asymmetric peg's orientation under occlusion. Grasp instability can also arise from the small contact patch between rounded Allegro fingertips and the rounded side of the peg during fast arm motion. Fingertip tactile sensing on Flexiv-Sharpa mitigates contact ambiguity, but robust object-centric perception under occlusion remains open. Wrist-mounted cameras and tactile sensing on additional platforms may help address these limitations.
\vspace{-2mm}

%% file: sections/5_discussion.tex
\section{Conclusion and Discussion}
\vspace{-2mm}



We presented ADEPT, an RL pre-training and RL post-training framework for learning dexterous arm-hand manipulation. A generic reposing task provides a reusable initialization for downstream manipulation, while structured post-training preserves useful pretrained behaviors as the policy adapts and aligns to new contact-rich task objectives. Vision and visuo-tactile student policies transfer these skills zero-shot to Kuka-Allegro and Flexiv-Sharpa for grasping, reorientation, transport, and insertion. Reusable pretraining reduces repeated skill discovery while supporting downstream refinement and new behaviors. Future work will broaden pretraining to tool use, clutter, and bimanual interaction.

%% file: sections/appendix.tex
\newpage
\appendix

\section{Task MDPs}

\subsection{Reposing Task}
\label{appx:reposingmdp}

The total reward is:

\begin{equation}
    r = \underbrace{w_r \cdot \exp(-\alpha_r \, d_h)}_{\text{hand-to-object alignment}} \;+\; \underbrace{w_g \cdot \exp(-\alpha_g \, e_{\mathrm{kp}}) \cdot g_c}_{\text{object-to-goal alignment}} \;+\; \underbrace{w_b \cdot \, g_c}_{\text{contact gating}}
\end{equation}

\noindent where:
\begin{itemize}[leftmargin=*]
\setlength\itemsep{2pt}
\item Reward component weights: $w_r = 1$, $w_g = 5$, $w_b = 0.01$.
\item $\alpha_r = 10$. $d_h = \max_j \| \mathbf{p}_j^{\mathrm{hand}} - \mathbf{p} \|_2$ is the maximum hand-to-object distance across all hand bodies.
\item $\alpha_g$: $15 \to 30$ (ADR). $e_{\mathrm{kp}}$ is the pose error, measured as the mean distance between 8 bounding-box keypoints of a cube of half-extent $h = 0.15$m of the current and goal object poses:
\begin{equation}
    e_{\mathrm{kp}} = \frac{1}{8} \sum_{i=1}^{8} \| \mathbf{K}_i(\mathbf{p}, \mathbf{q}) - \mathbf{K}_i(\mathbf{p}^*, \mathbf{q}^*) \|_2
\end{equation}
where $\mathbf{K}(\mathbf{p}, \mathbf{q}) \in \mathbb{R}^{8 \times 3}$ maps an object pose to its bounding-box corners in world frame.
\item $g_c \in \{0,1\}$ is a contact gate active when the thumb and at least one other finger exert force above threshold $\tau = 1$N.
\item $\mathbf{q}_h$ are hand joint positions and $\bar{\mathbf{q}}_h$ is a nominal uncurled configuration.
\item Gravity is annealed from $0$ to $-9.81$\,m/s$^2$ and object scale from $0.5\text{--}1.0\times$ over 50 ADR increments, advancing when success rate exceeds $0.4$. An episode succeeds when $e_{\mathrm{kp}} < 0.10$\,m. Episodes last 4\,s.

\end{itemize}

\subsection*{Domain Randomization (ADR)}

The following parameters are annealed from easy to hard over 50 ADR increments, triggered when the running success rate exceeds 0.4:

\begin{itemize}
    \item Gravity: $0 \to -9.81$\,m/s$^2$ (gravity annealing)
    \item Fabric damping gain: $10 \to 20$
    \item Goal sharpness: $-15 \to -30$ (tighter precision required)
\end{itemize}

\begin{figure}[h]
\centering
  \includegraphics[width=0.8\textwidth]{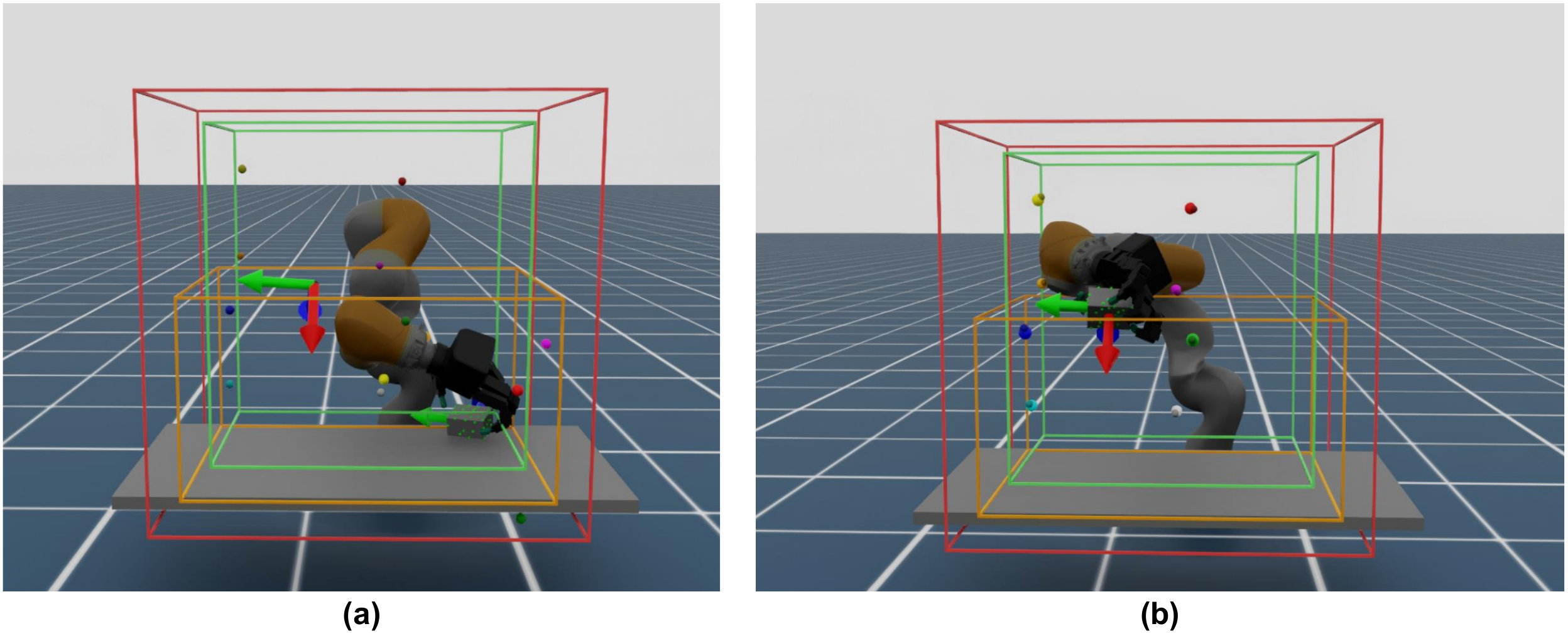}
  \caption{Reposing Task. Visualizes the out-of-bound range in red, the goal sampling range in green, and the object spawn range in orange. (a) visualizes a frame of the initial grasp of the object and (b) visualizes a frame where the robot has completed the task by reaching the target pose frame. The object point clouds can be seen in green spheres, which are on the object.}
\end{figure}

\subsection*{Primitive Objects for Pre-Training}

\begin{figure}[h]
\centering
  \includegraphics[width=1.0\textwidth]{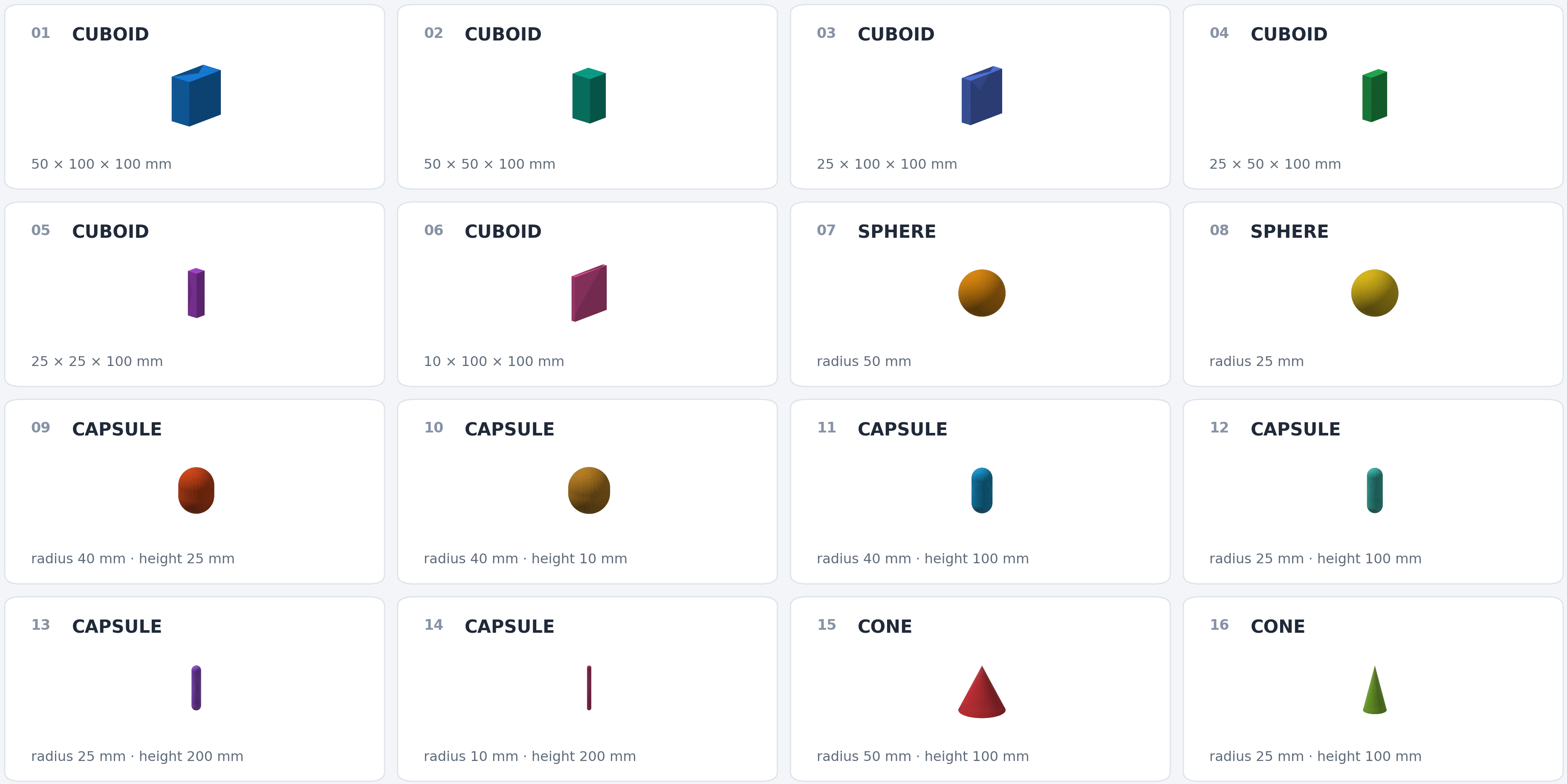}
  \caption{16 primitive objects used during pre-training reposing task. The scale and physical properties of the objects are randomized during pre-training.}
  \label{fig:primitive_objects}
\end{figure}

\subsection{Downstream Tasks from Scratch and Post-training}
\label{appx:downstreammdp}

All downstream manipulation tasks share a common reward function. Training from scratch and ADEPT post-training use the same reward parameters, with different initialization and training procedures.

The total reward is:
\begin{equation}
    r = \underbrace{w_r \cdot \exp(-\alpha_r \, d_h)}_{\text{hand-to-object alignment}} \;+\; \underbrace{w_g \cdot \exp(-\alpha_g \, e_{\mathrm{kp}})}_{\text{object-to-goal alignment}} 
\end{equation}

\noindent where:
\begin{itemize}[leftmargin=*]
\setlength\itemsep{2pt}
\item Both regimes use identical reward parameters: $w_r = 1$, $\alpha_r = 10$, $w_g = 5$, $\alpha_g = 15$ (constant), with the finger-curl and lift terms disabled ($w_c = w_\ell = 0$).
\item Goal tolerance anneals from $0.05 \to 0.02$\,m for both regimes over 50 ADR increments, advancing when success rate exceeds $0.4$. An episode succeeds when $e_{\mathrm{pos}} < \epsilon_{\mathrm{goal}}$.
\item $d_h = \max_j \| \mathbf{p}_j^{\mathrm{hand}} - \mathbf{p} \|_2$ is the maximum hand-to-object distance across all hand bodies.
\item $\mathbf{q}_h$ are hand joint positions and $\bar{\mathbf{q}}_h$ is a nominal uncurled configuration.
\item $\Delta z = |p_z - p_z^*|$ is the vertical distance between the object and its goal.
\item Both training settings use the following reward parameters:

\medskip
\begin{tabular}{@{}lcc@{}}
& \textbf{From scratch} & \textbf{Bootstrap} \\
$w_r$, $\alpha_r$ & $1$, $10$ & $1$, $10$ \\
$w_g$, $\alpha_g$ & $5$, $15$ (constant) & $5$, $15$ (constant) \\
\end{tabular}
\end{itemize}

\begin{figure}[h]
\centering
  \includegraphics[width=1.0\textwidth]{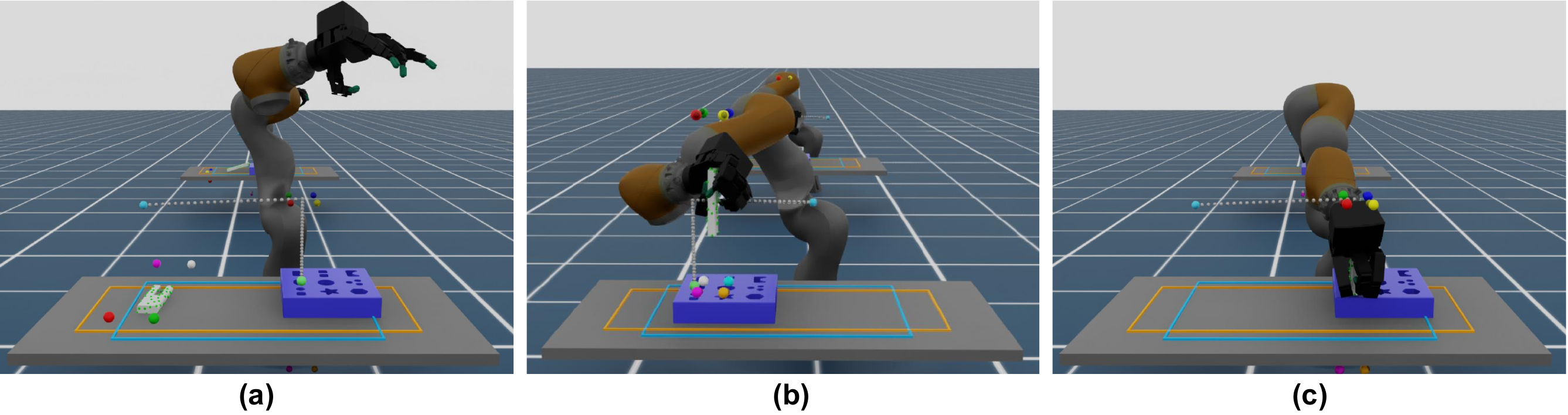}
  \caption{Downstream FMB Task. Visualizes the peg (orange) and board (sky blue) spawn ranges, object keypoints and goal pose keypoints. Grey keypoints represent the path set by the ADR level (denoted goal path annealing). At ADR 0, the goal is set to where the light blue sphere is placed and at ADR 50, the goal is set to the insertion goal where the light green sphere is placed. (a) shows sample initialization of the scene where the peg is dropped on to the table. In (b), we show how the reposing teacher reaches the goal at ADR 20, which is above the board. In (c), we visualize a sample policy rollout that inserts the peg into the board successfully where the keypoints align by the matching colors. The object point clouds in green spheres can be seen on the peg. }
  \label{fig:fmb_task_layout}
\end{figure}

\newpage
\subsection{ADEPT Pre-training and Post-training Algorithm}
\label{appx:bootstrapalgorithms}

Table~\ref{tab:training_hparams} compares the PPO hyperparameters used for reposing pretraining and FMB post-training. Pretraining uses a higher actor learning rate, looser PPO clip, and adaptive learning-rate schedule to encourage exploration of the broad reposing task; post-training reduces both substantially and switches to a linear learning-rate decay to keep the adapted policy close to the pretrained behavior. The critic learning rate is held constant at $5\times 10^{-5}$ across both stages, and the discount factor $\gamma$ and GAE $\tau$ are unchanged.

\begin{table}[h]
\centering
\small
\caption{PPO hyperparameter comparison between reposing pretraining and FMB post-training. The actor learning rate is reduced by $100\times$ and the PPO clip is tightened by $4\times$ during post-training, while the critic learning rate is held constant. The actor learning-rate schedule changes from KL-adaptive (pretraining, with target KL = Actor KL threshold) to linear decay over Max epochs (post-training); the Actor KL threshold is logged but not used during post-training under the linear schedule.}
\label{tab:training_hparams}
\setlength{\tabcolsep}{8pt}
\renewcommand{\arraystretch}{1.1}
\begin{tabular}{@{}l c c@{}}
\toprule
\textbf{Hyperparameter} & \textbf{Pretraining (reposing)} & \textbf{Post-training (FMB)} \\
\midrule
PPO Actor learning rate & $1 \times 10^{-3}$ & $1 \times 10^{-5}$ \\
Actor LR schedule & KL-adaptive & linear decay over Max epochs \\
BC actor learning rate & - & $1 \times 10^{-4}$ \\
Max epochs & $750{,}000$ & $200{,}000$ \\
Critic learning rate & $5 \times 10^{-5}$ & $5 \times 10^{-5}$ \\
PPO clip $\epsilon$ & $0.20$ & $0.05$ \\
Actor KL threshold & $0.010$ & --- \\
Critic KL threshold & $0.016$ & $0.016$ \\
Discount factor $\gamma$ & $0.998$ & $0.998$ \\
GAE $\tau$ & $0.95$ & $0.95$ \\
Mini-epochs (actor / critic) & $5 / 5$ & $4 / 4$ \\
Horizon length & $32$ & $16$ \\
Minibatch size & $16{,}384$ & $24{,}576$ \\
Entropy coefficient & $1 \times 10^{-3}$ & $2 \times 10^{-4}$ \\
Critic coefficient & $4$ & $4$ \\
Gradient-norm clip & $1.0$ & $1.0$ \\
Reward scale & $0.01$ & $0.01$ \\
\bottomrule
\end{tabular}
\end{table}

\begin{algorithm}[h]
\caption{Pretraining and Post-training for Bootstrapped Downstream Task Learning}
\label{alg:bootstrap}
\begin{algorithmic}[1]

\Statex \textbf{Input:} pretraining MDP $\mathcal{M}_{\mathrm{pre}}$ with obs $o^{\mathrm{pre}}$, post-training MDP $\mathcal{M}_{\mathrm{post}}$ with obs $o^{\mathrm{post}} \neq o^{\mathrm{pre}}$ (overlapping; neither is a superset), PPO clip $\epsilon$

\Statex
\Statex \textbf{Stage 1: reposing Teacher RL Training on $\mathcal{M}_{\mathrm{pre}}$}
\State Initialize pretrained policy $\pi_{\mathrm{pre}}(\cdot\,;\psi)$ and critic $V_{\mathrm{pre}}(\cdot\,;\phi_{\mathrm{pre}})$
\For{iteration $= 1, 2, \ldots$}
    \State Collect trajectories $\tau \sim \pi_{\mathrm{pre}}(\cdot \mid o_t^{\mathrm{pre}})$ in $\mathcal{M}_{\mathrm{pre}}$
    \State Update $\psi, \phi_{\mathrm{pre}}$ with PPO
\EndFor

\Statex
\Statex \textbf{Stage 2: Actor Warm-Start via BC Distillation}
\State Initialize downstream policy $\pi_{\mathrm{post}}(\cdot\,;\theta)$
\For{iteration $= 1, 2, \ldots$}
    \State Collect trajectories $\tau \sim \pi_{\mathrm{post}}(\cdot \mid o_t^{\mathrm{post}})$ in $\mathcal{M}_{\mathrm{post}}$
    \State Query teacher actions: $\mu_{\mathrm{pre}}, \sigma_{\mathrm{pre}} \gets \pi_{\mathrm{pre}}(\cdot \mid o_t^{\mathrm{pre}})$
    \State $\mathcal{L}_{\mathrm{BC}} \gets \sqrt{\sum_i \frac{(\mu_{\mathrm{post}}^i(o_t^{\mathrm{post}}) - \mu_{\mathrm{pre}}^i)^2}{\sigma_{\mathrm{pre}}^{i\,2}}} + \| \sigma_{\mathrm{post}}(o_t^{\mathrm{post}}) - \sigma_{\mathrm{pre}} \|_2$
    \State Update $\theta$ to minimize $\mathcal{L}_{\mathrm{BC}}$
\EndFor

\Statex
\Statex \textbf{Stage 3: Frozen Actor Critic Warm-Up}
\State Initialize critic $V_{\mathrm{post}}(\cdot\,;\phi)$, freeze $\theta$
\For{iteration $= 1, 2, \ldots$}
    \State Collect trajectories $\tau \sim \pi_{\mathrm{post}}(\cdot \mid o_t^{\mathrm{post}})$ in $\mathcal{M}_{\mathrm{post}}$, compute returns $\hat{R}_t$ via GAE
    \State Update $\phi$ to minimize $\mathbb{E}_t[(V_{\mathrm{post}}(s_t) - \hat{R}_t)^2]$
\EndFor

\Statex
\Statex \textbf{Stage 4: RL Post-Training on $\mathcal{M}_{\mathrm{post}}$}
\State Unfreeze $\theta$
\For{iteration $= 1, 2, \ldots$}
    \State Collect trajectories $\tau \sim \pi_{\mathrm{post}}(\cdot \mid o_t^{\mathrm{post}})$ in $\mathcal{M}_{\mathrm{post}}$, compute $\hat{R}_t$, $\hat{A}_t$
    \State Update $\theta, \phi$ with PPO (clip $\epsilon$)
\EndFor

\Statex
\Statex \textbf{Stage 5: Vision-Based Student Distillation}
\State Initialize vision policy $\pi_\omega$
\For{iteration $= 1, 2, \ldots$}
    \State Collect trajectories $\tau \sim \pi_\omega(\cdot \mid o_t^{\mathrm{img}})$ in $\mathcal{M}_{\mathrm{post}}$ with cameras
    \State Query teacher actions: $\mu_{\mathrm{post}}, \sigma_{\mathrm{post}} \gets \pi_{\mathrm{post}}(\cdot \mid o_t^{\mathrm{post}})$
    \State $\mathcal{L}_{\mathrm{DT}} \gets \sqrt{\sum_i \frac{(\mu_\omega^i(o_t^{\mathrm{img}}) - \mu_{\mathrm{post}}^i)^2}{\sigma_{\mathrm{post}}^{i\,2}}} + \| \sigma_\omega(o_t^{\mathrm{img}}) - \sigma_{\mathrm{post}} \|_2$
    \State Update $\omega$ to minimize $\mathcal{L}_{\mathrm{DT}}$
\EndFor

\Statex
\State \Return $\pi_\omega$
\end{algorithmic}
\end{algorithm}

\newpage
\subsection{Observations per stage.}
\label{appx:obstable}
\begin{description}[font=\bfseries]
\item[Stage 1: reposing pretraining.]
PPO from scratch. Teacher (reposing policy) has privileged state; actor policy uses
proprioceptive state + contacts + point cloud; critic adds privileged object state.
No receptacle-specific observations.

\item[Stage 2: Bootstrap to downstream task.]
BC warm-start + conservative PPO. Initializes from the Stage 1 reposing actor.
Adds receptacle pose and object--receptacle contact to both
actor and critic. Critic keeps Stage 1's privileged object state.

\item[Stage 3: Distillation (state $\to$ vision).]
BC from the Stage 2 state actor (now playing the teacher role). The new student
replaces the point cloud with two RGB camera images (left and center).
\end{description}

\begin{table}[H]
\centering
\small
\caption{Observation fields consumed at each pipeline stage in the current Cspace-mode codebase.
Stage~1 = reposing pretraining (PPO, actor + asymmetric critic); Stage~2 = FMB post-training
(PPO, actor + asymmetric critic); Stage~3 = two-camera RGB vision distillation, where a student
network is supervised by the Stage~2 actor (treated as the teacher) via behavior cloning. Actor
and student rows read the noisy variant of each proprioceptive field; the critic reads the
corresponding clean variant. Each cell shows the dimensionality of the field when it is
consumed at that stage; \xmark{} means the field is not consumed at that stage, and ``image''
denotes an image-tensor input. Rows newly introduced at
\colorbox{stage2row}{Stage~2 (FMB extras / nominal delta)} and
\colorbox{stage3row}{Stage~3 (two-camera RGB)} are highlighted.}
\label{tab:obs-layout-v2}
\setlength{\tabcolsep}{5pt}
\begin{tabular}{@{}l cc cc cc@{}}
\toprule
 & \multicolumn{2}{c}{\textbf{Stage 1}: reposing}
 & \multicolumn{2}{c}{\textcolor{stage2hdr}{\textbf{Stage 2}: FMB}}
 & \multicolumn{2}{c}{\textcolor{stage3hdr}{\textbf{Stage 3}: Distillation}} \\
\cmidrule(lr){2-3} \cmidrule(lr){4-5} \cmidrule(lr){6-7}
\textbf{Observation field} & Actor & Critic & Actor & Critic & Teacher (actor) & Student (actor) \\
\midrule
\multicolumn{7}{@{}l}{\textit{Proprioceptive / robot state}} \\
\midrule
\texttt{robot\_dof\_pos} (noisy) & 23 & 23 & 23 & 23 & 23 & 23 \\
\texttt{robot\_dof\_vel} (noisy) & 23 & 23 & 23 & 23 & 23 & 23 \\
\texttt{hand\_pos} (5 bodies $\times$ 3) & 15 & 15 & 15 & 15 & 15 & 15 \\
\texttt{hand\_vel} (5 bodies $\times$ 3) & 15 & 15 & 15 & 15 & 15 & 15 \\
\texttt{actions} & 23 & 23 & 23 & 23 & 23 & 23 \\
\texttt{fabric\_q}, \texttt{qd}, \texttt{qdd} (cspace, 3$\times$23)
                                             & 69 & 69 & 69 & 69 & 69 & 69 \\
\texttt{fingertip\_contacts} (5 $\times$ 3) & 15 & 15 & 15 & 15 & 15 & \xmark \\
\rowcolor{stage2row} nominal delta (joint + fingertip)
                                             & \xmark & \xmark & \xmark & 38 & \xmark & 38 \\
\midrule
\multicolumn{7}{@{}l}{\textit{Task / object state}} \\
\midrule
\texttt{object\_pos} (noisy) & 3 & \xmark & \xmark & \xmark & \xmark & \xmark \\
\texttt{object\_rot} (noisy) & 4 & \xmark & \xmark & \xmark & \xmark & \xmark \\
object state (pos + quat + vel) & \xmark & 13 & \xmark & 13 & \xmark & \xmark \\
\texttt{object\_goal} (position) & 3 & 3 & 3 & 3 & 3 & \xmark \\
\texttt{object\_goal\_quat} & 4 & 4 & 4 & 4 & 4 & \xmark \\
\texttt{multi\_object\_idx\_onehot} & 1 & 1 & \xmark & \xmark & \xmark & \xmark \\
\texttt{object\_scale} & 1 & 1 & \xmark & \xmark & \xmark & \xmark \\
\midrule
\multicolumn{7}{@{}l}{\textit{Point cloud / vision}} \\
\midrule
\texttt{pointcloud} (64 pts $\times$ 3) & 192 & 192 & 192 & \xmark & 192 & \xmark \\
\rowcolor{stage3row} RGB left camera ($H \times W \times 3$) & \xmark & \xmark & \xmark & \xmark & \xmark & image \\
\rowcolor{stage3row} RGB center camera ($H \times W \times 3$) & \xmark & \xmark & \xmark & \xmark & \xmark & image \\
\midrule
\multicolumn{7}{@{}l}{\textit{Receptacle (FMB-specific) extras}} \\
\midrule
\rowcolor{stage2row} receptacle pose & \xmark & \xmark & 7 & 7 & 7 & \xmark \\
\rowcolor{stage2row} object $\leftrightarrow$ receptacle contact force
                                             & \xmark & \xmark & 3 & 3 & 3 & \xmark \\
\midrule
\multicolumn{7}{@{}l}{\textit{Privileged sim-only signals (critic only)}} \\
\midrule
hand body ang.\ vel.\ (5 bodies $\times$ 3) & \xmark & 15 & \xmark & \xmark & \xmark & \xmark \\
palm force (first 3 of wrench) & \xmark & 3 & \xmark & \xmark & \xmark & \xmark \\
palm wrench (full 6-D) & \xmark & \xmark & \xmark & 6 & \xmark & \xmark \\
joint torque (measured) & \xmark & 23 & \xmark & 23 & \xmark & \xmark \\
\midrule
\multicolumn{1}{r}{\textbf{Flat dim total}} & \textbf{391} & \textbf{438} & \textbf{392} & \textbf{280} & \textbf{392} & 206 $+$ imgs \\
\bottomrule
\end{tabular}
\end{table}

\begin{table}[H]
\centering
\small
\caption{\textbf{Flexiv--Sharpa (29 DoF).} Observation fields consumed by the
Rizon--Sharpa Cspace pipeline. Stage~1 uses the reposing PBT actor and asymmetric
critic; Stage~2 uses the FMB bootstrap teacher; Stage~3 distills that state-based
teacher into the stereo RGB + five-fingertip TacMap student. The entries and
dimensions follow the checked-in agent configurations. ``Image'' denotes image-tensor input.}
\label{tab:obs-layout-rizon-sharpa}
\setlength{\tabcolsep}{4pt}
\begin{tabular}{@{}l cc cc cc@{}}
\toprule
 & \multicolumn{2}{c}{\textbf{Stage 1}: reposing}
 & \multicolumn{2}{c}{\textcolor{stage2hdr}{\textbf{Stage 2}: FMB}}
 & \multicolumn{2}{c}{\textcolor{stage3hdr}{\textbf{Stage 3}: Distillation}} \\
\cmidrule(lr){2-3} \cmidrule(lr){4-5} \cmidrule(lr){6-7}
\textbf{Observation field} & Actor & Critic & Actor & Critic & Teacher (actor) & Student (actor) \\
\midrule
\multicolumn{7}{@{}l}{\textit{Proprioceptive / robot state}} \\
\midrule
\texttt{proprio} (29 joints + 6 FK bodies) & 94 & 94 & 94 & 94 & 94 & 94 \\
\texttt{last\_actions} & 29 & 29 & 29 & 29 & 29 & \xmark \\
\texttt{fabric\_q}, \texttt{qd}, \texttt{qdd} ($3\times29$)
                                             & 87 & 87 & 87 & 87 & 87 & 87 \\
\texttt{contact\_forces\_obs} (5 fingers $\times$ 3)
                                             & 15 & 15 & 15 & 15 & 15 & \xmark \\
\texttt{contact\_force\_palm} & 3 & 3 & 3 & 3 & 3 & \xmark \\
\midrule
\multicolumn{7}{@{}l}{\textit{Task / object state}} \\
\midrule
\texttt{object\_pos\_noisy} & 3 & \xmark & \xmark & \xmark & \xmark & \xmark \\
\texttt{object\_rot\_noisy} & 4 & \xmark & \xmark & \xmark & \xmark & \xmark \\
\texttt{object\_state} (pos + quat + velocity)
                                             & \xmark & 13 & \xmark & 13 & \xmark & \xmark \\
\texttt{object\_goal} & 3 & 3 & 3 & 3 & 3 & \xmark \\
\texttt{object\_goal\_quat} & 4 & 4 & 4 & 4 & 4 & \xmark \\
\texttt{multi\_object\_idx\_onehot} & 1 & 1 & \xmark & 1 & \xmark & \xmark \\
\texttt{object\_scale} & 1 & 1 & \xmark & 1 & \xmark & \xmark \\
\midrule
\multicolumn{7}{@{}l}{\textit{Point cloud / RGB / tactile perception}} \\
\midrule
\texttt{pc\_obs} (64 points $\times$ 3) & 192 & 192 & 192 & 192 & 192 & \xmark \\
\rowcolor{stage3row} stereo RGB (left + right) & \xmark & \xmark & \xmark & \xmark & \xmark & 2 images \\
\rowcolor{stage3row} \texttt{sharpa\_tactile\_deform}
                                             & \xmark & \xmark & \xmark & \xmark & \xmark & 5 images \\
\rowcolor{stage3row} \texttt{sharpa\_tactile\_anchor\_xyz} ($5\times3$)
                                             & \xmark & \xmark & \xmark & \xmark & \xmark & 15 \\
\midrule
\multicolumn{7}{@{}l}{\textit{FMB receptacle extras}} \\
\midrule
\rowcolor{stage2row} \texttt{board\_pose} & \xmark & \xmark & 7 & 7 & 7 & \xmark \\
\rowcolor{stage2row} \texttt{object\_target\_contact\_force}
                                             & \xmark & \xmark & 3 & 3 & 3 & \xmark \\
\midrule
\multicolumn{7}{@{}l}{\textit{Privileged sim-only critic signals}} \\
\midrule
\texttt{hand\_forces} (palm wrench) & \xmark & 6 & \xmark & 6 & \xmark & \xmark \\
\texttt{measured\_joint\_torque} & \xmark & 29 & \xmark & 29 & \xmark & \xmark \\
\midrule
\multicolumn{1}{r}{\textbf{Flat dim total}} & \textbf{436} & \textbf{477} & \textbf{437} & \textbf{487} & \textbf{437} & 196 + images \\
\bottomrule
\end{tabular}
\end{table}
\newpage

\section{Joint Configuration Space Geometric Fabric Setup}
\label{appx:fabric_setup}

This appendix details the geometric fabric components, parameters, and
policy--fabric integration loop summarized in Sec.~\ref{sec:cspace_fabrics}.
We adopt the notation of \cite{dextrahg}: the fabric state $\mathbf{q}_f, \dot{\mathbf{q}}_f, \ddot{\mathbf{q}}_f \in \mathbb{R}^{n_q}$ evolves according to
\begin{equation}
\label{eq:fabric_ode}
\mathbf{M}_f(\mathbf{q}_f, \dot{\mathbf{q}}_f)\, \ddot{\mathbf{q}}_f
\;+\; \mathbf{f}_f(\mathbf{q}_f, \dot{\mathbf{q}}_f)
\;+\; \mathbf{f}_\pi(\mathbf{a}_t) \;=\; \mathbf{0},
\end{equation}
where $\mathbf{M}_f \succ 0$ is the system metric (mass) assembled from all
fabric components by pull-back, $\mathbf{f}_f$ is the nominal path-generating
geometric force, and $\mathbf{f}_\pi$ is the action-driven forcing term.
The policy emits a cspace target $\mathbf{q}^*_t = \boldsymbol{\pi}(\mathbf{a}_t)
\in \mathbb{R}^{n_q}$ at $60\,\text{Hz}$; the fabric is forward-integrated with
the second-order Runge--Kutta scheme of \cite{dextrahg} at $60\,\text{Hz}$.
For the 23 DoF KUKA-Allegro system ($n_q = 7\;\text{arm} + 16\;\text{finger}$) and the 29 DoF Flexiv-Sharpa system ($n_q = 7\;\text{arm} + 22\;\text{finger}$),
we partition the cspace target into arm and hand slices,
$\mathbf{q}^*_t = [\mathbf{q}^{*,a}_t;\,\mathbf{q}^{*,h}_t]$ with $\mathbf{q}^{*,a}_t
\in \mathbb{R}^7$ and $\mathbf{q}^{*,h}_t \in \mathbb{R}^{16}$ or $\mathbf{q}^{*,h}_t \in \mathbb{R}^{22}$ respectively.

\subsection{Pull-Back from Taskmaps to the Root}

Every fabric component lives on a taskmap $\mathbf{x} =
\boldsymbol{\phi}(\mathbf{q}_f)$ with Jacobian
$\mathbf{J} = \partial \boldsymbol{\phi} / \partial \mathbf{q}_f$ and
contributes a leaf metric $\mathbf{M}(\mathbf{x},\dot{\mathbf{x}})$ and leaf
force $\mathbf{f}(\mathbf{x},\dot{\mathbf{x}})$. Root-space contributions are
\begin{equation}
\label{eq:pullback}
\mathbf{M}_f \;\mathrel{+}=\; \mathbf{J}^\top \mathbf{M}\, \mathbf{J}, \qquad
\mathbf{f}_f \;\mathrel{+}=\; \mathbf{J}^\top \!\left(\mathbf{f}
\;+\; \mathbf{M}\, \dot{\mathbf{J}}\, \dot{\mathbf{q}}_f \right),
\end{equation}
the second term being the curvature force required to keep the leaf
acceleration equal to the pulled-back leaf force under the chain rule.

\subsection{Cspace Attractors}
\label{appx:cspace_attractor}

For each body part $p \in \{a,h\}$ (arm and hand), the cspace taskmap is the
slice $\mathbf{x}^p = \mathbf{S}_p \mathbf{q}_f$ with $\mathbf{S}_p$ the
constant index-selector. The position error is $\mathbf{e}^p =
\mathbf{x}^p - \mathbf{q}^{*,p}_t$.

\paragraph{Mass.} The metric is isotropic with a smooth switch between a
minimum and maximum mass around the error norm:
\begin{equation}
\label{eq:attr_mass}
m^p(\mathbf{e}^p) \;=\; (m^p_{\max} - m^p_{\min})\,
\sigma\!\left(-\sigma_m^p\big(\|\mathbf{e}^p\| - \delta_m^p\big)\right)
\;+\; m^p_{\min},
\qquad
\mathbf{M}^p(\mathbf{e}^p) \;=\; m^p(\mathbf{e}^p)\, \mathbf{I},
\end{equation}
where $\sigma(z) = \tfrac{1}{2}(\tanh z + 1)$, $\sigma_m^p$ is the
\emph{mass sharpness} and $\delta_m^p$ the \emph{mass switch offset}. By default, $m^p_{\min} = m^p_{\max}$, so the
switch is inert and $\mathbf{M}^p = m^p \mathbf{I}$.

\paragraph{Forcing attractor.} A homogeneous-degree-one (HD1) acceleration
pulls $\mathbf{x}^p$ toward $\mathbf{q}^{*,p}_t$ with conical shaping along the
error direction and a soft attraction-radius damping:
\begin{align}
\ddot{\mathbf{x}}^p_{\text{forc}}
&= -\,k^p_a\, \tanh\!\big(\alpha^p_a \|\mathbf{e}^p\|\big)\,
\frac{\mathbf{e}^p}{\|\mathbf{e}^p\|}
\;-\; b^p(\mathbf{e}^p)\, \dot{\mathbf{x}}^p,
\label{eq:attr_forcing} \\[2pt]
b^p(\mathbf{e}^p)
&= b^p_{\max}\, \sigma\!\left(-\sigma_d^p \big(\|\mathbf{e}^p\| - r_d^p\big)\right),
\label{eq:attr_damping}
\end{align}
with conical gain $k^p_a$, conical sharpness $\alpha^p_a$, damping ceiling
$b^p_{\max}$, damping sharpness $\sigma_d^p$, and damping radius $r_d^p$. The
damping engages only inside the attraction radius $r_d^p$, providing critical
damping near the target without slowing long approaches.

\paragraph{Geometric attractor.} In parallel with the forcing term we attach a
homogeneous-degree-two (HD2) geometric attractor with the same direction but
quadratic-in-velocity scaling
\begin{equation}
\label{eq:attr_geom}
\ddot{\mathbf{x}}^p_{\text{geom}}
\;=\; -\,k^{p,g}_a\, \|\dot{\mathbf{x}}^p\|^{2}\,
\tanh\!\big(\alpha^{p,g}_a \|\mathbf{e}^p_g\|\big)\,
\frac{\mathbf{e}^p_g}{\|\mathbf{e}^p_g\|},
\end{equation}
where $\mathbf{e}^p_g = \mathbf{x}^p - \mathbf{q}^{p}_{\mathrm{post}}$ and $\mathbf{q}^{p}_{\mathrm{post}}$ is a config-driven default that postures the elbow out and fingers curled. Since $\ddot{\mathbf{x}}^p_{\text{geom}}$
is HD2 in $\dot{\mathbf{x}}^p$, it produces speed-invariant paths and does
not bias convergence to $\mathbf{q}^{*,p}_t$ in steady state \cite{dextrahg}.

\subsection{Body Sphere Collision Avoidance}
\label{appx:collision}

We approximate the robot geometry with $N_s = 31$ collision spheres covering
the arm links, palm, knuckles, and finger phalanges. Forward kinematics maps
$\mathbf{q}_f$ to sphere origins $\mathbf{x}^{(i)}_s = \boldsymbol{\phi}_{\mathrm{fk},i}(\mathbf{q}_f)$
with Jacobian $\mathbf{J}^{(i)}_s \in \mathbb{R}^{3 \times n_q}$. For each
sphere--obstacle pair (environment meshes and declared self-collision
sphere--sphere pairs), let $r_i$ be the sphere radius, $d_i$ the signed
clearance, $\hat{\mathbf{n}}_i = (\mathbf{r}_i - \mathbf{x}^{(i)}_s) /
\|\mathbf{r}_i - \mathbf{x}^{(i)}_s\|$ the unit direction to the nearest
obstacle point $\mathbf{r}_i$, and $\dot{\mathbf{x}}^{(i)}_s$ the sphere
velocity. The signed closing speed is
\begin{equation}
v_i \;=\; -\,\hat{\mathbf{n}}_i \cdot (\dot{\mathbf{x}}^{(i)}_s - \mathbf{v}^{\mathrm{obs}}_i),
\end{equation}
with $\mathbf{v}^{\mathrm{obs}}_i$ the velocity of the closest point on the
obstacle (rigid composition of linear and angular velocity, supporting moving
obstacles). Following \cite{dextrahg}, define
\begin{align}
s_i &\;=\; \tfrac{1}{2}\!\left(\tanh\!\big(-\alpha_1(v_i - \alpha_2)\big) + 1\right),
\qquad \text{(smooth velocity gate)}
\label{eq:velgate} \\
\bar{d}_i &\;=\; \max(d_{\min},\, d_i),
\end{align}
where $s_i$ vanishes when the sphere retreats and saturates to $1$ when it
approaches. The leaf metric and leaf acceleration for sphere $i$ are
\begin{equation}
\label{eq:coll_leaf}
\mathbf{M}^{(i)}_b \;=\; \frac{s_i}{\bar{d}_i^{\,3}}\;\hat{\mathbf{n}}_i \otimes \hat{\mathbf{n}}_i,
\qquad
\ddot{\mathbf{x}}^{(i)}_b \;=\; \frac{s_i}{\bar{d}_i^{\,3}}\;\hat{\mathbf{n}}_i.
\end{equation}
A per-sphere Frobenius normalization preserves Eigen-directions while making
the response invariant to absolute distance,
$\widehat{\mathbf{M}}^{(i)}_b = \mathbf{M}^{(i)}_b / \|\mathbf{M}^{(i)}_b\|_F$,
and the resulting block is rescaled by a budget-allocated weight that caps
the aggregate stiffness across all spheres,
\begin{equation}
\label{eq:budget}
w_i \;=\; \frac{\kappa\, (r_i / \|\mathbf{r}\|_2)}{\bar{d}_i^{\,2}}, \qquad
\widetilde{w}_i \;=\; \frac{\min\!\big(\sum_j w_j,\; \mathcal{B}\big)\, w_i}{\sum_j w_j},
\qquad
\widetilde{\mathbf{M}}^{(i)}_b \;=\; \widetilde{w}_i\, \widehat{\mathbf{M}}^{(i)}_b,
\end{equation}
where $\kappa$ is a metric scalar and $\mathcal{B}$ is the total metric
budget (separate values for forcing vs.\ geometric channels). The leaf
acceleration is split into a forcing and a geometric component:
\begin{align}
\ddot{\mathbf{x}}^{(i)}_{b,\text{forc}}
&\;=\; -\,k_f\, \hat{\mathbf{n}}_i \;-\; b_c\,
(\dot{\mathbf{x}}^{(i)}_s - \mathbf{v}^{\mathrm{obs}}_i),
\label{eq:coll_forcing} \\
\ddot{\mathbf{x}}^{(i)}_{b,\text{geom}}
&\;=\; -\,k_g\, \|\dot{\mathbf{x}}^{(i)}_s - \mathbf{v}^{\mathrm{obs}}_i\|^{2}\, \hat{\mathbf{n}}_i.
\label{eq:coll_geom}
\end{align}
Each is pulled back through its own sphere Jacobian and summed:
\begin{equation}
\mathbf{M}^{\mathrm{coll}}_f
\;=\; \sum_i \mathbf{J}^{(i)\top}_s\, \widetilde{\mathbf{M}}^{(i)}_b\, \mathbf{J}^{(i)}_s,
\qquad
\mathbf{f}^{\mathrm{coll}}_f
\;=\; -\sum_i \mathbf{J}^{(i)\top}_s\, \widetilde{\mathbf{M}}^{(i)}_b\, \ddot{\mathbf{x}}^{(i)}_b.
\end{equation}
Per-sphere pull-back (rather than stacked-vector pull-back) ensures that the
budget weighting acts on the unit-Frobenius blocks $\widehat{\mathbf{M}}^{(i)}_b$
rather than on the cross-sphere normalization, eliminating dilution as $N_s$
grows.

\subsection{Joint Limit Repulsion}
\label{appx:jl}

Two taskmaps, one per limit side, are defined for each joint $i$:
\begin{equation}
x^{\,U}_i \;=\; \frac{\bar{q}_i - q_{f,i}}{\bar{q}_i - \underline{q}_i},
\qquad
x^{\,L}_i \;=\; \frac{q_{f,i} - \underline{q}_i}{\bar{q}_i - \underline{q}_i},
\qquad x^{\,U}_i, x^{\,L}_i \in [0,1],
\end{equation}
i.e. the signed clearance is normalized by the joint range, so all tuning
parameters live in ``fraction of range'' units regardless of physical units
(rad vs.\ m for revolute vs.\ prismatic joints). The Jacobian rows are
$\pm \mathrm{diag}\big(1/(\bar{q}_i - \underline{q}_i)\big)$. The leaf metric
and acceleration on each side are
\begin{align}
\label{eq:jl_metric}
\mathbf{M}^{\,J}(\mathbf{x}, \dot{\mathbf{x}})
&\;=\; \mathrm{diag}\!\left(\,g_i(\dot{x}_i)\, \frac{k_b}{(\max(x_i - \delta_J,\, \epsilon))^{2}}\,\right),
\\[2pt]
g_i(\dot{x}_i) &\;=\; \sigma\!\left(-\sigma_g(\dot{x}_i - \delta_g)\right) \in [0,1], \\
\ddot{\mathbf{x}}^{\,J}_{\text{forc}}
&\;=\; \mathbf{g}\,k_r \;-\; b_J\,\mathbf{1}\{\dot{\mathbf{x}} < 0\}\odot\dot{\mathbf{x}}, \\
\ddot{\mathbf{x}}^{\,J}_{\text{geom}}
&\;=\; \|\dot{\mathbf{x}}\|^{2}\, k_r\, \mathbf{g},
\label{eq:jl_accel}
\end{align}
where $k_b$ is the metric scalar, $\delta_J$ the metric-exploder offset (the
barrier saturates within the last $\delta_J$ of the range), $\epsilon$
clamps the metric below a configured maximum, $k_r$ is the soft-ReLU
repulsion gain, and $b_J$ a damping gain that engages only when $\dot{x}_i < 0$. The
gate $g_i$ is a smooth tanh switch (replacing the bang-bang gate of
\cite{dextrahg}) that tapers the barrier when the joint retreats from the
limit, suppressing chatter at the approach/retreat boundary while preserving
the unbounded $1/(x{-}\delta_J)^2$ stiffness as $x \to \delta_J$.

\paragraph{Per-joint acceleration and jerk caps.} On top of the JL barrier, a
post-hoc quadratic-program step \cite{dextrahg} rescales the integrated
acceleration to respect per-joint caps,
\begin{equation}
\ddot{q}_{f,i}^{\,\max} \;=\; \min\!\left(\ddot{q}_i^{\,\max},\;
\Delta t \cdot \frac{\dddot{q}_i^{\,\max}}{2\, \ddot{q}_i^{\,\max}}\right),
\end{equation}
with per-joint values
$\ddot{q}^{\,\max}_a = \{7.5,\,7.5,\,10,\,10,\,10,\,20,\,20\}\,\text{rad/s}^2$
for the arm and $\ddot{q}^{\,\max}_h = 22.5\,\text{rad/s}^2$ for all sixteen
hand joints (jerk caps scale accordingly).

\subsection{Cspace Damping and Speed Control}
\label{appx:speed}

Two velocity-dependent terms regulate end-to-end motion smoothness and speed.
The first is a constant cspace damping applied to the
mass-momentum $\mathbf{M}_f \dot{\mathbf{q}}_f$:
\begin{equation}
\label{eq:cspace_damping}
\mathbf{f}^{\,\mathrm{damp}}_f \;=\; k_d\, \mathbf{M}_f\, \dot{\mathbf{q}}_f,
\qquad k_d = 10 \;\;\text{(exposed to ADR).}
\end{equation}
The second is a speed-control term that caps the full system kinetic-energy level $E^\star$. We attach
Euclidean energies to two taskmaps:
(i) a slice $\mathbf{x}^{\overline{h}} = \mathbf{S}_{\overline{h}} \mathbf{q}_f$
over the non-hand joints (here, the 7 arm joints) with scale
$\beta_{\overline{h}} = 0.75$, and (ii) a robot-frame-origins taskmap
$\mathbf{x}^{\mathrm{palm}} \in \mathbb{R}^3$ over the palm origin with
scale $\beta_{\mathrm{palm}} = 0.25$. Each contributes
\begin{equation}
\mathbf{M}_E^{(k)} = \beta_k\, \mathbf{I}, \qquad
E^{(k)} \;=\; \tfrac{1}{2}\, \beta_k\, \dot{\mathbf{x}}^{(k)\top} \dot{\mathbf{x}}^{(k)},
\qquad k \in \{\overline{h}, \mathrm{palm}\}.
\end{equation}
Hand joints are deliberately excluded so finger motion is decoupled from the
energization budget (finger joints are free to move as fast as fabric terms allow). The pulled-back energy metric and total energy are
\begin{equation}
\mathbf{M}_E \;=\; \sum_k \mathbf{J}^{(k)\top}\, \mathbf{M}_E^{(k)}\, \mathbf{J}^{(k)},
\qquad
E(\mathbf{q}_f, \dot{\mathbf{q}}_f) \;=\; \sum_k E^{(k)}.
\end{equation}
The energization coefficient \cite{dextrahg} that holds $E$ constant under
the current geometric force is
\begin{equation}
\label{eq:alpha}
\alpha \;=\; -\,
\frac{\dot{\mathbf{q}}_f^{\,\top}\!\left(\mathbf{M}_E\, \ddot{\mathbf{q}}_{f,\mathrm{geom}}
+ \mathbf{f}_E\right)}{\dot{\mathbf{q}}_f^{\,\top}\, \mathbf{M}_E\, \dot{\mathbf{q}}_f + \varepsilon},
\qquad
\ddot{\mathbf{q}}_{f,\mathrm{geom}} \;=\; -\mathbf{M}_f^{-1}\, \mathbf{f}_f,
\end{equation}
with $\mathbf{f}_E = \mathbf{0}$ for Euclidean energies. The energized force
is $\mathbf{f}_f^\star = \mathbf{f}_f - \alpha\, \mathbf{M}_f\, \dot{\mathbf{q}}_f$,
which annihilates the velocity component along the geometric flow. Finally,
the speed-control damping engages only when the current energy exceeds the
target:
\begin{equation}
\label{eq:speed_control}
\mathbf{f}^{\,\mathrm{sc}}_f \;=\; b_{\mathrm{sc}}\, \mathbf{1}\{E > E^\star\}
\;\mathbf{M}_f\, \dot{\mathbf{q}}_f,
\qquad
E^\star = 1.0,\;\; b_{\mathrm{sc}} = 100 \;\;\text{(both exposed to ADR).}
\end{equation}
Equations~\eqref{eq:cspace_damping}--\eqref{eq:speed_control} yield a
trajectory whose end-to-end speed is consistent across configurations: long
motions are energized to the target, short motions are passively damped.

\subsection{Action-to-Cspace Mapping}

The policy outputs $\mathbf{a}_t \in [-1, 1]^{n_q}$ with $n_q = 23$ for KUKA-Allegro ($7$ arm + $16$ finger joints) and $n_q = 29$ for Flexiv-Sharpa ($7$ arm + $22$ finger joints), interpreted as per-joint relative deltas. For KUKA-Allegro, the delta scale is $\delta_{\max} = 0.1$\,rad. The Cspace target that enters the policy forcing term $\mathbf{f}_\pi$ is $\mathbf{q}^*_t = \mathbf{q}_f^{(t-1)} + \delta_{\max}\,\mathbf{a}_t$, clamped to the URDF joint limits, where $\mathbf{q}_f^{(t-1)}$ is the fabric's integrated joint configuration from the previous step. This relative formulation keeps each policy step bounded in magnitude and avoids the large position jumps that an absolute target would allow, while still letting the policy exploit the full $n_q$-dimensional action manifold over the course of a trajectory.

\subsection{Policy--Fabric Integration Loop}

At each control step, the action-mapped Cspace target $\mathbf{q}^*_t$ is broadcast to the fabric. The fabric performs $K$ inner integration steps of the fabric ODE with timestep $\Delta t$ via a displacement integrator (default $K = 2$), producing $(\mathbf{q}_f, \dot{\mathbf{q}}_f, \ddot{\mathbf{q}}_f)$. The integrated position and velocity are then issued to the underlying joint controller (PD in simulation, an admittance controller on the real robot). The fabric is captured into a CUDA graph so the per-step cost remains negligible at the parallel scale required for large-batch RL.

\subsection{ADR-Annealed Fabric Parameters}

Several fabric parameters are part of the ADR curriculum rather than fixed at YAML defaults:
\begin{itemize}[leftmargin=*]
\setlength\itemsep{2pt}
\item \textbf{Cspace damping gain.} Annealed from a high value (heavily smoothed, easy to control) toward a lower value (more reactive) as success rate improves.
\item \textbf{Speed-control energy target.} Annealed to broaden the range of motion speeds the policy is exposed to.
\item \textbf{Velocity attenuation.} A scalar applied between the fabric velocity output and the joint controller, also annealed via ADR.
\end{itemize}
The fabric is therefore not treated as a fixed black-box controller but as part of the curriculum: as policy competence grows, the regularization the fabric provides is gradually relaxed.

\section{Population-Based Training}
\label{appx:pbt}

We train the reposing pretraining teacher with decentralized Population-Based Training (PBT)~\cite{pbt} to efficiently search the joint PPO + ADR hyperparameter space.

\textbf{Aggregate simulated experience.} For Flexiv-Sharpa, approximately $100$ hours of pretraining at $75{,}000$ environment transitions per second per node across $16$ nodes yields $4.32\times10^{11}$ transitions. At a $60$\,Hz control rate, this corresponds to approximately $228$ years of aggregate simulated robot experience:
\begin{equation}
\frac{100\times3600\times75{,}000\times16}{60\times(365\times24\times3600)} \approx 228.3\;\text{years}.
\end{equation}
The estimate sums experience across the training population. Using the same aggregate throughput of $1.2\times10^6$ transitions per second, Kuka-Allegro pretraining over about $25$ hours corresponds to approximately $57$ years of aggregate simulated robot experience:
\begin{equation}
\frac{25\times3600\times1.2\times10^6}{60\times(365\times24\times3600)} \approx 57.1\;\text{years}.
\end{equation}

\textbf{Ranking and replacement.} We launch $N = 16$ policies in parallel. After a warm-up of $T_{\mathrm{start}} = 200\text{M}$ environment frames, every policy independently performs a PBT check every $T_{\mathrm{PBT}} = 200\text{M}$ frames: it reads the latest metadata files from all other policies, restricts the comparison to checkpoints from policies that have collected approximately the same number of frames (so newer policies are not penalized for having less data), and ranks the population by target objective. The bottom $r_{\mathrm{replace}} = 0.4$ fraction of the population is eligible for replacement: an eligible policy within a small fractional-standard-deviation and absolute threshold of the population leader only has its hyperparameters mutated, while one significantly behind the leader additionally has its policy weights overwritten with those of a uniformly-randomly-selected top-$r_{\mathrm{replace}}$ policy.

\textbf{ADR is preserved on replacement.} When a low-performing policy's weights are rewritten from a top performer, it continues training from its own ADR level rather than the donor's. Because the receiving policy typically has a lower ADR level than the donor, the rewritten policy effectively resumes training against an easier curriculum slice but with the donor's well-shaped weights, letting it climb back toward the population frontier without re-discovering the basic dexterous behaviors the donor already encodes.

\textbf{Mutation.} At every PBT check, each of the following PPO hyperparameters is independently re-sampled with probability $p_{\mathrm{mut}} = 0.25$ via a log-uniform perturbation (multiplied or divided by a factor sampled uniformly in $[1.1, 2.0]$):
\begin{itemize}[leftmargin=*]
\setlength\itemsep{2pt}
\item Learning rate, gradient-norm clip, entropy coefficient, critic coefficient, bounds-loss coefficient, and KL threshold (all via \texttt{mutate\_float}).
\item PPO clip $\epsilon$ via \texttt{mutate\_eps\_clip}, clamped to $[0.01, 0.3]$.
\item Number of PPO mini-epochs via \texttt{mutate\_mini\_epochs}, clamped to $[1, 12]$ with $\pm 1$--$3$ integer steps.
\item Discount factor $\gamma$ and GAE $\tau$ via \texttt{mutate\_discount}, which mutates $1 - x$ rather than $x$ to be conservative on values close to $1$.
\end{itemize}

\textbf{Target objective.} Policies are ranked by their running success rate at their current ADR level, exposed to PBT via the simulator's \texttt{true\_objective} info field rather than the shaped reward. Using the unshaped success rate as the ranking signal means PBT compares policies on actual task progress rather than on shaped-reward magnitudes that fluctuate with mutated PPO hyperparameters (e.g.\ entropy coefficient, value loss scale) and with each policy's ADR level.


\section{Post-Training Recipes}
\label{appx:post-training-recipes}

This appendix details Stages~2--4 of the structured post-training procedure (Alg.~\ref{alg:bootstrap}, Sec.~\ref{sec:posttraining}). PPO hyperparameters are listed in Tab.~\ref{tab:training_hparams}.

\textbf{Stage 2 --- Actor warm-start via BC distillation.}
\begin{itemize}[leftmargin=*]
\setlength\itemsep{2pt}
\item BC for $40{,}000$ iterations is sufficient to converge the student's action distribution to the teacher's on the downstream observation space.
\item BC rollouts use a mixed student-teacher policy that alternates between teacher and student actions, which keeps the BC dataset on the states the student will visit at deployment while remaining anchored on teacher-reachable states.
\item The critic is not updated during BC; only the actor minimizes the Mahalanobis-weighted distance to the teacher action distribution.
\item BC is pinned at ADR step 20 (Sec.~\ref{sec:sim-exp}), where the pretrained teacher zero-shots reliably: the goal is the single ADR 20 pose on the lift-and-transport segment, and ADR advancement is disabled during BC. The goal-path annealing flag is kept on so the student tracks the teacher through the lift-and-transport segment.
\end{itemize}

\textbf{Stage 3 --- Critic warm-up with frozen actor.}
\begin{itemize}[leftmargin=*]
\setlength\itemsep{2pt}
\item From this stage onward, goal-path annealing is turned off and the goal is set to the final insertion endpoint at ADR 50; the ADR randomizations continue to advance. For the first $20$ post-BC epochs the actor is frozen and only the critic updates against $r_{\mathrm{post}}$. This aligns value estimates with $\mathcal{M}_{\mathrm{post}}$ before any policy-gradient step modifies the actor.
\item During this window the actor's log-standard-deviation is fixed at $\log\sigma = -2$ ($\approx 0.14$ in linear scale), keeping exploration narrow around the BC-distilled mean so the critic targets stay on-distribution.
\item The critic learning rate is $5\times 10^{-5}$ throughout (unchanged from Stage 4).
\end{itemize}

\textbf{Stage 4 --- Conservative PPO post-training.}
\begin{itemize}[leftmargin=*]
\setlength\itemsep{2pt}
\item Actor learning rate drops to $1\times 10^{-5}$ ($100\times$ lower than pretraining) with linear decay over $200{,}000$ epochs; PPO clip ratio tightens to $\epsilon = 0.05$ ($4\times$ tighter) to discourage large policy updates.
\item Goal-path annealing is turned off in PPO: the goal sits at the actual insertion endpoint so PPO learns the contact-rich insertion itself, while ADR can continue to advance.
\item The post-training actor--critic uses separate actor and critic trunks (rather than the shared-trunk architecture of pretraining), allowing the critic to specialize on $r_{\mathrm{post}}$ without contaminating actor features.
\item Observation clipping is relaxed from $5.0$ (default) to $100.0$ to accommodate the larger magnitudes of receptacle pose and object--receptacle contact-force observations introduced at post-training.
\end{itemize}

\textbf{Task-specific reward shaping for FMB peg insertion.}
\begin{itemize}[leftmargin=*]
\setlength\itemsep{2pt}
\item Both peg--board contact penalty and receptacle contact penalty are set to $0$. The pretrained policy already handles contacts gracefully, and insertion explicitly requires sustained contact between the peg, fingertips, and receptacle.
\item External object wrenches are disabled during post-training, since the peg is grasped throughout.
\item Peg orientation randomization is set to its maximum range ($[1.0, 1.0]$) from the start of post-training rather than ramped via ADR, and the object-to-goal reward sharpness is fixed at $15$ rather than annealed.
\end{itemize}

\section{ADEPT Training Efficiency}
\label{appx:training_efficiency}

The reported training-curve runs use four NVIDIA L40S GPUs each. The from-scratch comparison includes ten independent seeds, while the component ablations use five independent non-PBT seeds per variant. PBT is used only for reposing pretraining. The reported 8B-step pretraining budget counts one PBT member, rather than aggregate experience across the population. We train a separate teacher and student for each peg.

\begin{figure}[H]
  \centering
  \includegraphics[width=0.7\textwidth]{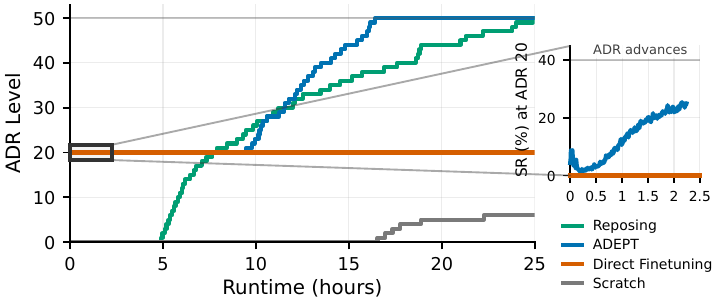}
  \caption{Training curves on FMB peg insertion comparing \textbf{(1)} training from scratch, \textbf{(2)} direct PPO finetuning of the pretrained reposing teacher, and \textbf{(3)} ADEPT post-training. Blue inset: success-rate collapse for (2) during transfer.}
  \label{fig:post-training-wallclock}
\end{figure}

\begin{figure}[H]
  \centering
  \includegraphics[width=0.7\textwidth]{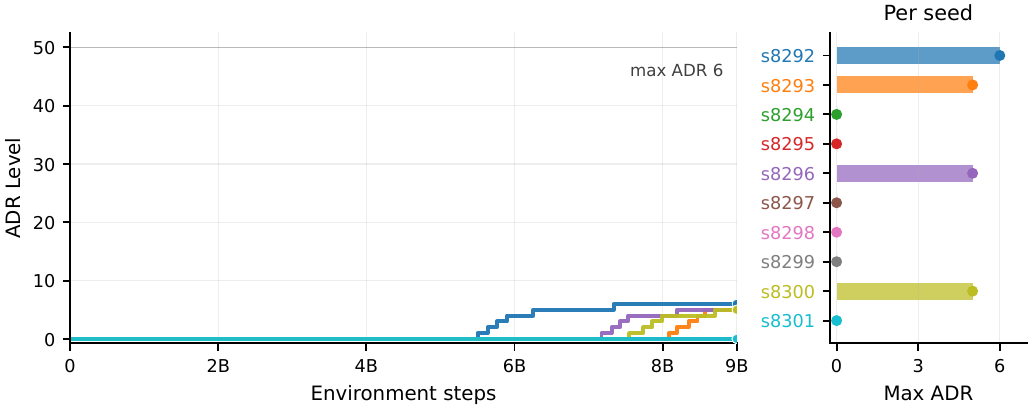}
  \caption{Training curves across different seeds for training from scratch. ADEPT post-training reliably converges to a solution while most seeds for training from scratch either plateau or take a very long time to train for a fixed compute budget.}
  \label{fig:across-seeds}
\end{figure}

\section{Real-World Deployment Topology}
\label{appx:deployment}

\begin{figure}[H]
  \centering
  \includegraphics[width=1.0\textwidth]{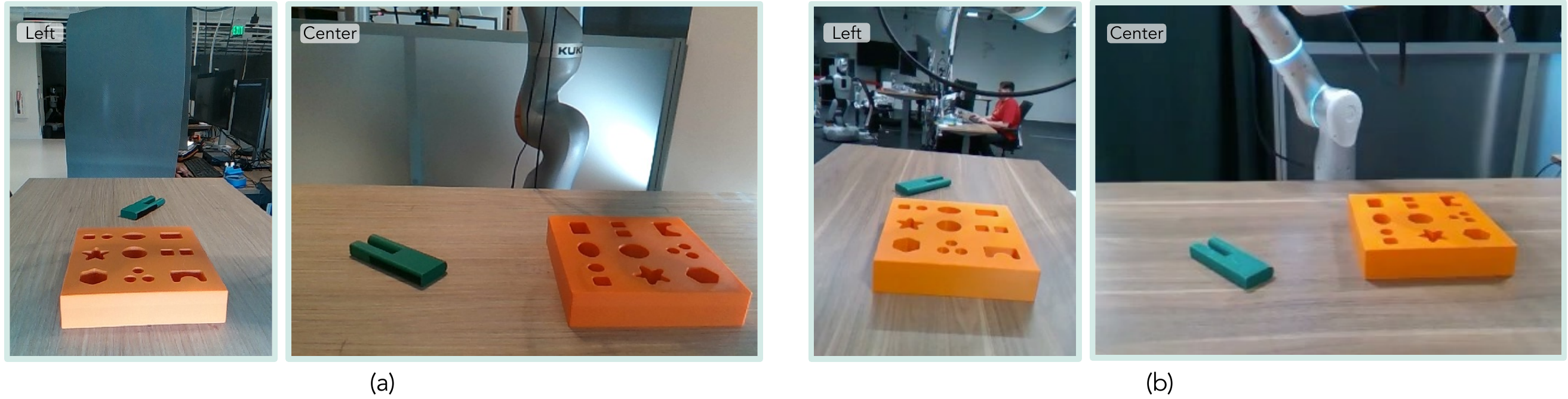}
  \caption{Real-world camera views for (a) Kuka-Allegro and (b) Flexiv-Sharpa setups.}
  \label{fig:camera_views}
\end{figure}

The KUKA-Allegro deployment described here uses the same Cspace action interface and observation layout as in simulation. The C++ admittance controller and RealSense driver provide physical control and camera observations in place of the simulator and renderer.

\textbf{Camera transport.} The two Intel RealSense cameras (one mounted to the left of the workspace, one centered above the workbench) are read by a C++ camera driver that resizes the images on-board to $320 \times 240$ and publishes them to the policy process over ZMQ.

\textbf{Compute split and processes.} The control stack is split across two machines. An NVIDIA Jetson AGX Orin sits next to the robot and hosts (i) a real-time C++ admittance controller that integrates the fabric-derived mass-and-force pair $(\mathbf{M}_f, \mathbf{f}_f + \mathbf{f}_\pi)$ from the fabric ODE in Sec.~\ref{sec:cspace_fabrics} into joint position and velocity commands, and (ii) the Python fabric process that evaluates the geometric fabric of Sec.~\ref{sec:cspace_fabrics}. The student policy runs on a separate workstation with a discrete GPU and receives proprioception and the two RGB streams over ZMQ; it issues 23-D Cspace targets back to the fabric, which integrates them into the smooth joint trajectories consumed by the admittance controller. Joint state and the $(\mathbf{M}_f, \mathbf{f}_f + \mathbf{f}_\pi)$ pair are exchanged between the C++ controller and the fabric process through POSIX shared memory on the Jetson.

\textbf{Rate hierarchy.} The C++ admittance controller runs at 1\,kHz, with the underlying KUKA control loop at 1\,kHz and the Allegro control loop at 333\,Hz. All Python-side processes (fabric, sensor, policy) share a common $60$\,Hz cadence, while the high-rate hardware loops run underneath.

\section{Distillation Loss Details}
\label{appx:distill_loss}

This appendix gives the full forms of the behavior-cloning and auxiliary losses that constitute the distillation objective $\mathcal{L} = \mathcal{L}_{\mathrm{BC}} + \mathcal{L}_{\mathrm{aux}}$ introduced in Sec.~\ref{sec:distillation}. The student policy $\pi_\theta(a_t \mid o_t)$ is a stochastic Gaussian with mean $\mu_\theta(o_t)$ and standard deviation $\sigma_\theta(o_t)$, and the teacher $\pi_T(a_t \mid o_t^T)$ has mean $\mu_T(o_t^T)$ and fixed standard deviation $\sigma_T$.

\subsection{Behavior-Cloning Loss}

The BC term matches both moments of the student's action distribution to the teacher's via a Mahalanobis-weighted L2 distance,
\begin{equation}
    \mathcal{L}_{\mathrm{BC}}(o_t, o_t^T;\,\theta) = \sqrt{\sum_{i=1}^{n_q} \frac{\bigl(\mu_\theta^i(o_t) - \mu_T^i(o_t^T)\bigr)^2}{(\sigma_T^i)^2}} \;+\; \sqrt{\sum_{i=1}^{n_q} \frac{\bigl(\sigma_\theta^i(o_t) - \sigma_T^i\bigr)^2}{(\sigma_T^i)^2}},
\end{equation}
where $n_q = 23$ for KUKA-Allegro and $n_q = 29$ for Flexiv-Sharpa. The teacher's fixed std-dev $\sigma_T$ serves as a per-dimension precision weighting: joints the teacher is more confident about (smaller $\sigma_T$) contribute more strongly to the loss.

\subsection{Auxiliary 8-Keypoint Pose Loss}

The auxiliary head predicts the eight bounding-box corners of the peg in the world frame, and the loss is the L2 norm of the per-corner residuals,
\begin{equation}
    \mathcal{L}_{\mathrm{aux}}(o_t;\,\theta) = \sqrt{\sum_{i=1}^{8} \bigl\| \hat{\mathbf{k}}^i_\theta(o_t) - \mathbf{k}^{\star,i}_t \bigr\|_2^2},
\end{equation}
where $\hat{\mathbf{k}}^i_\theta(o_t) \in \mathbb{R}^3$ is the student's predicted 3D world-frame position of the $i$-th corner, and $\mathbf{k}^{\star,i}_t \in \mathbb{R}^3$ is the corresponding ground-truth corner position obtained by rotating the eight half-extent offsets of the peg by the simulator's peg orientation and adding the peg position. 

\subsection{Total Distillation Objective}

Combining the two terms with their respective weighting coefficients,
\begin{equation}
    \mathcal{L}_{\mathrm{distill}}(\theta) = \mathbb{E}_{o_t \sim \pi_\theta}\!\left[\, w_{\mathrm{BC}} \cdot \mathcal{L}_{\mathrm{BC}}(o_t, o_t^T;\,\theta) \;+\; w_{\mathrm{aux}} \cdot \mathcal{L}_{\mathrm{aux}}(o_t;\,\theta) \,\right],
\end{equation}
where $w_{\mathrm{BC}}$ and $w_{\mathrm{aux}}$ are scalar weighting coefficients and the expectation is over rollouts $o_t \sim \pi_\theta$ collected by the student in the same task environment. We use $w_{\mathrm{BC}}=1$ and $w_{\mathrm{aux}}=20$.

\section{Visuo-Tactile Student Architecture}
\label{appx:visuotactile_arch}

This appendix details the visuo-tactile student used on the Flexiv-Sharpa embodiment. The vision-only KUKA-Allegro student is a strict subset of the same architecture with the tactile branch removed.

\textbf{Observations.} The student consumes the two-camera RGB pair $(\mathbf{I}^L_t, \mathbf{I}^C_t)$ from left and center cameras, noisy proprioception $\mathbf{q}_t$, fabric state $\mathbf{q}^f_t$, per-finger TacMap depth maps $\{\mathbf{T}^k_t\}_{k=1}^{5}$, and per-finger fingertip positions $\{\mathbf{x}^k_t\}_{k=1}^{5} \in \mathbb{R}^3$ obtained via forward kinematics on the noisy joint state.

\textbf{Vision encoder.} The two-camera pair is passed through a shared ResNet backbone followed by a cross-attention fuser~\cite{dextrahrgb} to a $d_{\text{vis}}$-dimensional vision latent ($d_{\text{vis}} = 256$). The backbone is initialized from the Stage-1 pretrained student (Sec.~\ref{sec:distillation}) and unfrozen on the DextrAH-RGB~\cite{dextrahrgb} schedule.

\textbf{Tactile encoder.} Each per-finger depth map $\mathbf{T}^k_t \in \mathbb{R}^{H\times W}$ is stacked with its thresholded binary contact map $\mathbb{1}[\mathbf{T}^k_t \geq \tau]$ ($\tau = 1/255$) into a two-channel input, and passed through a small CNN (two $3{\times}3$ convolutions with stride 2 and channel widths 16 then 32, adaptive average pooling, and a linear projection with LayerNorm and ELU) shared across fingers. The CNN emits a per-finger feature $\mathbf{u}^k_t \in \mathbb{R}^{d_f}$ with $d_f = 32$.

\textbf{FiLM spatial anchoring.} Each fingertip position $\mathbf{x}^k_t$ is expanded with $B{=}4$ Fourier bands, $\gamma(\mathbf{x}) = [\mathbf{x},\, \sin(2^0 \pi \mathbf{x}), \cos(2^0 \pi \mathbf{x}), \dots, \sin(2^{B-1} \pi \mathbf{x}), \cos(2^{B-1} \pi \mathbf{x})]$, and mapped through a two-layer MLP (hidden 128, ELU) to per-finger FiLM parameters $(\boldsymbol{\gamma}^k_t, \boldsymbol{\beta}^k_t) \in \mathbb{R}^{d_f} \times \mathbb{R}^{d_f}$. The modulated per-finger feature is
\begin{equation}
    \tilde{\mathbf{u}}^k_t = (1 + \alpha \boldsymbol{\gamma}^k_t) \odot \mathbf{u}^k_t + \alpha \boldsymbol{\beta}^k_t,
\end{equation}
with $\alpha = 0.1$ scaling the modulation to keep it a soft perturbation on the shared per-finger CNN features rather than a hard override. The five modulated per-finger features are flattened into a tactile latent $\mathbf{u}_t \in \mathbb{R}^{d_{\text{tac}}}$ with $d_{\text{tac}} = 5 \cdot d_f = 160$.

\textbf{Fusion and policy trunk.} The vision, tactile, and proprioception latents are concatenated with the fabric state and passed through a two-layer MLP with widths $[512, 512]$ and ELU activations, followed by a 1024-unit LSTM (layer-normalized, placed before the final linear heads). The final head emits a mean action $\boldsymbol{\mu}_\theta(o_t) \in \mathbb{R}^{n_q}$ with a fixed log-standard-deviation, matching the teacher's action parameterization.

\textbf{Training objective.} The student is trained with pure teacher BC using the objective in Appx.~\ref{appx:distill_loss}, with a single auxiliary term supervising an 8-keypoint peg pose prediction from the shared vision features. On Flexiv-Sharpa FMB the aux term uses a peg-tight-keypoints target with coefficient $w_{\mathrm{aux}} = 20$; other auxiliary heads (object position, object 6-D rotation, board pose, per-finger contact force and binary contact) are exposed by the codebase but disabled for this experiment. The BC target is filtered through a soft z-mask that down-weights teacher actions in low-height regions of the state distribution (threshold $0.08$ m, softness $0.02$, floor $0.1$) to avoid over-fitting to teacher table-clearance behavior that the real robot cannot exactly reproduce.




\section{Distillation Domain Randomization}
\label{appx:distill_randomization}

To enable zero-shot sim-to-real transfer, both stages of the teacher--student distillation curriculum (Sec.~\ref{sec:distillation}) apply aggressive domain randomization across physics, scene layout, sensor noise, and visual rendering. The randomization schedule for each stage matches the corresponding RL training stage so that the student observes the same environmental distribution that the teacher was trained on. Table~\ref{tab:distill_randomization} summarizes the full set of parameters and their ranges; the values reported correspond to the Stage~2 (downstream FMB) regime, with the analogous Stage~1 ranges differing only in scene layout (single fixed receptacle, full $\mathrm{SO}(3)$ peg orientation).

\begin{figure}[H]
\centering
  \includegraphics[width=\textwidth]{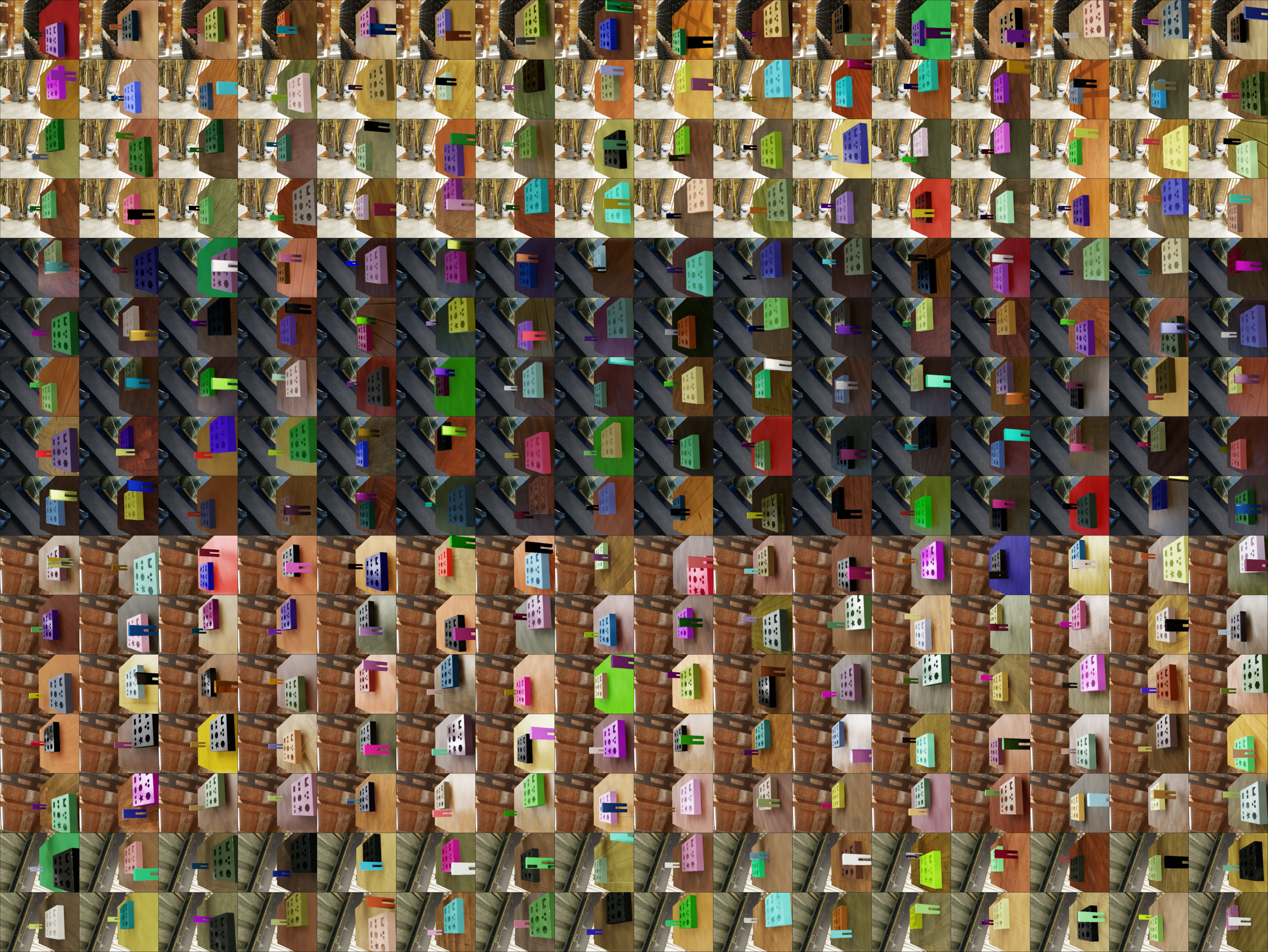}
  \caption{A sample 16x16 grid of visual randomizations on the left camera of the FMB training scene for distillation.}
\end{figure}

\newpage

\begin{table}[H]
\centering
\small
\caption{Domain randomization parameters used during teacher--student distillation.
Per-env ranges are sampled uniformly at episode reset; ADR-annealed ranges advance
from the lower bound to the upper bound over training as the running success rate
exceeds the ADR threshold; per-step noise is added at every control step; per-frame
randomization is applied at each camera render.}
\label{tab:distill_randomization}
\setlength{\tabcolsep}{8pt}
\renewcommand{\arraystretch}{1.1}
\begin{tabular}{@{}l l@{}}
\toprule
\textbf{Parameter} & \textbf{Range / Distribution} \\
\midrule
\multicolumn{2}{@{}l}{\textit{Physics (per-env at reset, ADR-annealed)}} \\
\midrule
Object mass scale & $(0.5,\,3.0) \times$ nominal \\
Object static friction & $(0.5,\,1.2)$ \\
Object dynamic friction & $(0.3,\,1.0)$ \\
Object restitution & $(0.8,\,1.0)$ \\
Robot joint stiffness scale & $(0.5,\,2.0) \times$ default \\
Robot joint damping scale & $(0.5,\,2.0) \times$ default \\
Robot joint friction & $\mathcal{U}(0,\,5.0)$\,Nm \\
\midrule
\multicolumn{2}{@{}l}{\textit{Scene and object spawn (per-env at reset, ADR-annealed)}} \\
\midrule
Peg spawn position $(x, y)$ jitter & up to $\pm 15$\,cm per axis \\
Peg spawn orientation & full $\mathrm{SO}(3)$ at final ADR \\
Random wrench on peg & up to $10$\,m/s$^2$ linear acceleration equivalent \\
Receptacle / board pose jitter & max range $[-7,12]$ cm in $x$, $[-30, 10]$ cm in $y$ \\
\midrule
\multicolumn{2}{@{}l}{\textit{Proprioception and object-state noise (per-step on observations)}} \\
\midrule
Joint position Gaussian noise & $\mathcal{N}(0,\,0.08^2)$\,rad \\
Joint position bias (per-env) & $\mathcal{U}(-0.08,\,0.08)$\,rad \\
Joint velocity Gaussian noise & $\mathcal{N}(0,\,0.18^2)$\,rad/s \\
Joint velocity bias (per-env) & $\mathcal{U}(-0.08,\,0.08)$\,rad/s \\
Object position Gaussian noise & $\mathcal{N}(0,\,0.03^2)$\,m \\
Object orientation Gaussian noise & $\mathcal{N}(0,\,0.1^2)$\,rad \\
\midrule
\multicolumn{2}{@{}l}{\textit{Visual randomization (per-frame at camera rendering)}} \\
\midrule
Lighting (intensity, color, direction) & random per env \\
Dome / background texture & random from texture library \\
Table material & random color and roughness \\
Robot link materials & random color and roughness \\
Peg diffuse tint (RGB) & per-channel $\mathcal{U}(0.90,\,1.0)$ (near-white) \\
Board diffuse tint (RGB) & per-channel $\mathcal{U}(0.04,\,0.12)$ (near-black) \\
Camera position jitter & $\pm 3$\,cm from calibrated pose \\
Camera rotation jitter & $\pm 3^\circ$ from calibrated pose \\
\bottomrule
\end{tabular}
\end{table}

\section{Qualitative Analysis of Emergent Grasping Behaviors}
\label{appx:qualitative_grasps}

\begin{figure}[h]
\centering
\includegraphics[width=1.0\textwidth]{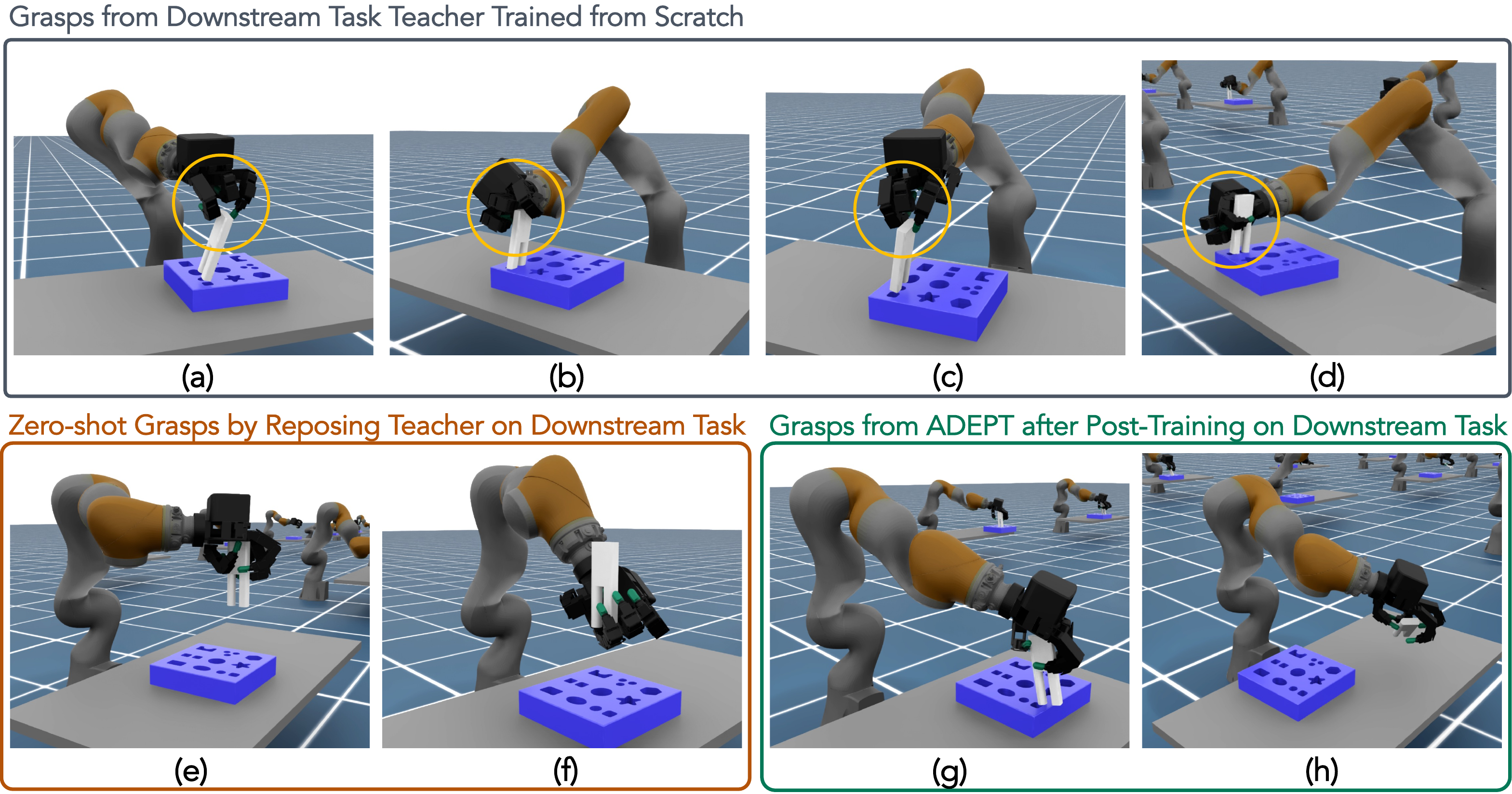}
\caption{Comparison of grasp behaviors. (a, b, c, d) Downstream-task teacher trained from scratch at ADR level 50, with grasps that lack the natural finger gaits emerging from reposing pretraining. (e, f) Pretrained reposing teacher applied zero-shot to FMB peg insertion at ADR level 20, occasionally producing undesirable grasps (e.g., bottom-end, upside-down) that solve the reposing segment but fail at insertion.  (g, h) Downstream-task teacher trained with ADEPT at ADR level 50, with natural grasps where the fingers wrap around and cage the peg.}
\label{fig:grasp_comparisons}
\end{figure}

Fig.~\ref{fig:grasp_comparisons} highlights representative grasp modes from each training condition. Panels (e, f) show the reposing teacher applied zero-shot at ADR level 20: the grasps look natural, with the fingers wrapping around the peg in multi-contact closure, but they are not always task-aligned. A common failure mode is grasping the peg bottom-up with the hand below the peg as shown in (f). This works for the reposing segment but cannot complete insertion, which requires grasping the peg from above with its legs pointing down toward the hole. Panels (g) and (h) show that ADEPT post-training removes the bottom-up grasps entirely while preserving the natural finger-wrap behavior inherited from reposing pretraining. Panels (a-d) show that policies trained from scratch do not develop the natural finger-wrap behavior at all; the grasps are functional for insertion but lack the multi-contact stability that reposing pretraining produces. For example, (a) and (c) grasp the peg using only the index and pinky fingers, a configuration that is highly unusual for precise human manipulation. In (b) and (d), the fingers are curled to their extreme with just one finger wrapping around the peg to stably grasp yet this is also very unusual and unnatural compared to the emergent grasp qualities displayed by the policies that have pre-training knowledge on manipulating diverse objects on a generic reposing task objective like in (e-h).  

